\documentclass{article}

\usepackage{arxiv}

\renewcommand{\shorttitle}{KungfuAthleteBot: Learning High-Dynamic Humanoid Motion}

\usepackage[utf8]{inputenc}
\usepackage[T1]{fontenc}
\usepackage{textcomp}
\usepackage{amsmath}
\usepackage{amsfonts}
\usepackage{graphicx}
\usepackage{xcolor}
\usepackage{booktabs}
\usepackage{multirow}
\usepackage{subcaption}
\usepackage{bbding}
\usepackage{makecell}
\usepackage{microtype}
\usepackage{hyperref}
\usepackage{url}
\usepackage{natbib}

\providecommand{\Description}[1]{}

\newcommand{\calQ}{\mathcal{Q}}
\newcommand{\calB}{\mathcal{B}}
\newcommand{\tick}{\textmd{\textcolor{green}{\Checkmark}}}
\newcommand{\cross}{\textmd{\textcolor{red}{\XSolidBrush}}}
\newcommand{\projectpage}{https://kungfuathletebot.github.io/}

\title{KungfuAthleteBot: Learning High-Dynamic Humanoid Motion from Video\\ with Unified Robust Recovery}

\author{
Zhongxiang Lei$^{1}$ \quad
Lulu Cao$^{1,2}$ \quad
Xuyang Wang$^{1,2}$ \quad
Tianyi Qian$^{3\dagger}$ \quad
Jinyan Liu$^{1\dagger}$ \quad
Xuesong Li$^{1\dagger}$ \\
\\
$^{1}$ Beijing Institute of Technology \\
$^{2}$ Equal contribution \\
$^{3}$ QIYUAN Lab \\
$^{\dagger}$ Corresponding authors
}

\hypersetup{
pdftitle={KungfuAthleteBot: Learning High-Dynamic Humanoid Motion from Video  with Unified Robust Recovery},
pdfsubject={cs.RO, cs.AI, cs.LG},
pdfauthor={Zhongxiang Lei, Lulu Cao, Xuyang Wang, Tianyi Qian, Jinyan Liu, Xuesong Li},
pdfkeywords={Humanoid Robots, Learning from Video, Motion Tracking, Reinforcement Learning, Fall Recovery, Video-based Motion Reconstruction},
}

\begin{document}
\maketitle

\begin{abstract}
Video is an abundant, inexpensive source of human motion data that is rich in extreme athletic behaviours. Making it usable for humanoid robots, however, is not a matter of simply retargeting a reconstructed trajectory: video-derived motion is \emph{physically inconsistent}, \emph{devoid of actuation information}, and \emph{says nothing about failure or recovery}. We present KungfuAthleteBot (KAB), a framework that treats ``learning high-dynamic motion from video'' as the central problem and resolves each of these three failure modes in turn. (C1) We build the KungfuAthlete dataset from videos of national-level martial artists and introduce a physics-guided parabolic trajectory correction that removes height floating, ground penetration, and high-frequency jitter from reconstructed aerial and landing phases. (C2) Because video carries no force information, strict tracking of a reconstructed trajectory is dynamically infeasible, and error-driven initialisation keeps re-launching the policy from infeasible aerial poses. We introduce physics-driven \emph{pseudo-low-kinetic-energy (LKE) sampling}, our central mechanism for making such references learnable: it biases initialisation towards dynamically feasible states, letting the policy discover feasible actuation patterns instead of imitating infeasible ones. (C3) Finally, we introduce a direct training paradigm in which disturbance rejection and fall recovery are learned inside the \emph{same} policy that tracks the video motion, requiring no recovery reference data and no manual mode switching. On a real humanoid robot, KAB learns highly dynamic skills from video and recovers from arbitrary falls in $\approx 0.7$\,s, to our knowledge the fastest reported recovery for a unified policy. Ablations on the unified policy confirm the necessity of its components, supporting the view that repairing and compensating video data -- rather than only collecting more of it -- is what unlocks high-dynamic humanoid skills. Code and data: \url{\projectpage}.
\end{abstract}


\maketitle

\section{Introduction}

\begin{figure}[t]
    \centering
    \includegraphics[width=1\linewidth]{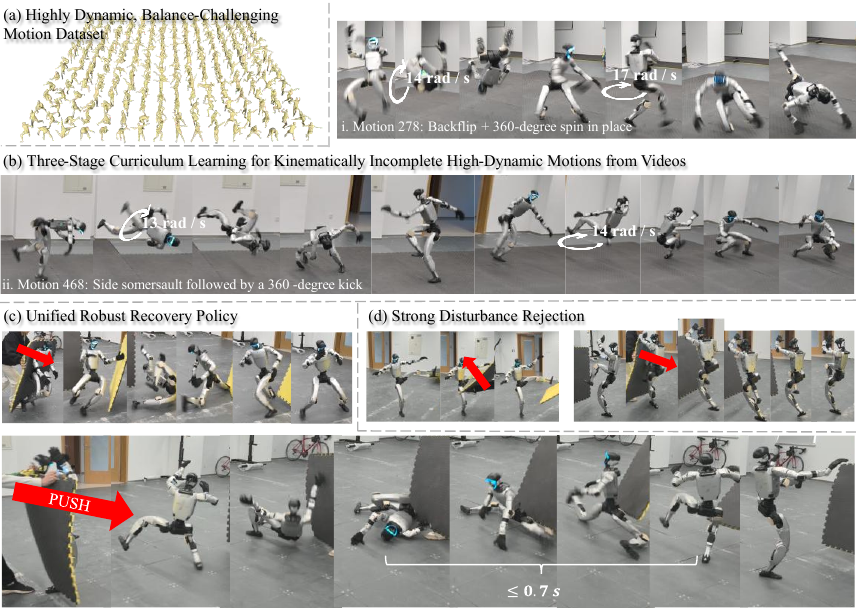}
    \caption{Learning high-dynamic motion from video. (a) A dataset extracted from videos of national-level martial artists. (b) Video-derived motion carries no actuation information, so direct mimic-based tracking (e.g., BeyondMimic) fails on such references. Our curriculum (Sec.~\ref{sec:curriculum}) and low-kinetic-energy sampling (Sec.~\ref{sec:plke}) let the policy explore feasible actuation patterns. (c) A single policy performs fall recovery and motion tracking, and (d) rejects external disturbances.}
    \label{fig:introduction}
    \Description{Overview of the video-to-robot pipeline: a dataset extracted from martial-arts videos, three-stage curriculum learning for high-dynamic tracking, unified fall recovery, and disturbance rejection on a humanoid robot.}
\end{figure}

Humanoid robots that can execute highly dynamic, athletic skills -- aerial spins, flips, and explosive strikes -- are a long-standing goal of the field (Fig.~\ref{fig:introduction}). The dominant bottleneck is no longer the control algorithm but the \emph{data} that drives it~\cite{liaoBeyondMimicMotionTracking2025,wang2026omnixtreme,sunRobotDancingResidualActionReinforcement2025}. Motion-capture pipelines require instrumented studios, wearable hardware, and bespoke collection procedures: they are expensive, slow to scale, and restricted to controlled environments. This paper starts from a different premise: compared with motion capture and wearable systems, video is abundant, inexpensive, and able to capture elite athleticism outside the laboratory.

Learning high-dynamic motion from video, however, is harder than learning from curated motion-capture data, and this is the problem we address. Retargeting a reconstructed human trajectory to a humanoid and running reinforcement learning on it does not work out of the box, for reasons that are specific and diagnosable:

\begin{itemize}
    \item \textbf{(C1) Video reconstructions are physically inconsistent.} Monocular reconstruction methods such as GVHMR~\cite{shenWorldGroundedHumanMotion2024} and generative models such as GENMO~\cite{li2025genmo} produce motion that is globally smooth but locally unphysical: during flight the root height ``floats'' instead of following a ballistic arc, the feet penetrate the ground on landing, and frame-wise noise appears as joint jitter. This is not a cosmetic issue -- a floating root makes the aerial phase dynamically impossible for any robot.
    \item \textbf{(C2) Video contains no actuation information.} A trajectory reconstructed from pixels describes \emph{where the body goes}, not \emph{how the joints are driven} to get there. In explosive motions such as a backflip, the human and the robot realise the same joint trajectory with radically different torque profiles (e.g.\ at the knee); demanding strict tracking therefore asks the robot to reproduce a physically infeasible joint trajectory, and learning stalls.
    \item \textbf{(C3) Video says nothing about failure.} A video ends when the motion ends; it provides no model of disturbance, loss of balance, or recovery. A policy that faithfully tracks video motion is therefore also a policy that has no idea what to do once the real world pushes it off the reference trajectory.
\end{itemize}

Our central claim is that these three failure modes are not incidental implementation details but the \emph{actual} obstacles to learning high-dynamic motion from video, and that they can only be overcome jointly. Accordingly, the paper is organised as a single pipeline, where each technical contribution is precisely the remedy for one failure mode, and where an ablation shows that removing any one of them breaks the pipeline.

\textbf{The video-to-robot pathway.} Videos are segmented into clips, reconstructed to a kinematic sequence $\mathbf{q}^{\mathrm{hum}}_{1:T}$, retargeted~\cite{araujoRetargetingMattersGeneral2025} to the robot to give a reference $\mathbf{q}^{\mathrm{ref}}_{1:T}$, and tracked by a policy $\pi$. The last three steps are where the pathway breaks, and they require two sub-systems: a \emph{repair} stage that restores physical validity, and a \emph{compensation} stage that supplies, through learning, the dynamics and the failure model that video never contained. These are not independently swappable: without the compensation, strict single-stage tracking of an infeasible reference collapses (Fig.~\ref{fig:exp_tracking}a), and without the unified recovery the robot cannot return to the motion once disturbed (Sec.~\ref{sec:ablations}).

\textbf{Contributions.}
\begin{itemize}
    \item \textbf{A video-derived professional high-dynamic dataset (addresses C1).} In collaboration with national-level martial artists, we construct KungfuAthlete from 197 training videos, segmented into 1{,}726 sub-clips, reconstructed and retargeted to a Unitree G1. It covers both sustained ground skills and highly non-stationary aerial skills, with higher body linear and angular velocities than the corpora compared in Table~\ref{tab:motion_stats}. Dataset, licence and released artefacts are detailed in Sec.~\ref{sec:dataset_release}.
    \item \textbf{A physics-aware repair of video artifacts (addresses C1).} We replace the floating root trajectory with a physically consistent one: ground alignment from the lowest body point, event-aware propagation with contact detection, ballistic (constant-acceleration) interpolation of flight phases, penetration removal, and Savitzky--Golay smoothing. Unlike prior work that simply \emph{discards} jump segments as unusable, our correction \emph{recovers} them, preventing a large-scale waste of the most valuable video data.
    \item \textbf{Physics-driven sampling for missing actuation (addresses C2).} We introduce \emph{pseudo-low-kinetic-energy (LKE) sampling}, our central mechanism for making unreliable video references learnable: rather than initialising wherever the current tracking error is largest, it derives the initial-state distribution from a kinetic-energy proxy of the reference and therefore never concentrates capacity on dynamically infeasible aerial poses. A standard progressive-tolerance curriculum complements it (Sec.~\ref{sec:curriculum}); Sec.~\ref{sec:curriculum_relation} contrasts our setting -- unreliable video-derived references -- with prior high-dynamic tracking work that assumes faithful references.
    \item \textbf{A unified robust policy over video data (addresses C3).} We propose a direct formulation in which one policy tracks the video motion, rejects disturbances, and recovers from arbitrary gravity-induced falls, with no recovery-specific reference data and no manual switching. It attains $\approx 0.7$\,s recovery -- to our knowledge the fastest reported for a unified policy -- validated on real hardware.
\end{itemize}

Sec.~\ref{sec:related} positions this work against prior art. The remainder of the paper is organised around the three failure modes: Sec.~\ref{sec:dataset} presents the dataset and the repair of reconstructed motion (C1); Sec.~\ref{sec:compensate} addresses missing actuation (C2); Sec.~\ref{sec:unified} presents the unified policy (C3); Sec.~\ref{sec:experiments} evaluates tracking, recovery and ablations, with additional material in the appendix.

\section{Related Work}
\label{sec:related}

\paragraph{Data and capture.}
Humanoid motion tracking has largely been driven by motion-capture corpora such as AMASS~\cite{mahmoodAMASSArchiveMotion2019} and LAFAN1~\cite{harvey2020robust}, which are natural and continuous but confined to low-to-medium dynamics, and by hardware-based collection such as exoskeletons (HOMIE, HumanoidExo)~\cite{ben2025homie,zhongHumanoidExoScalableWholeBody} or VR teleoperation (TWIST2, SONIC)~\cite{ze2025twist2,luo2025sonic}, which are physically consistent but require specialist hardware and scale poorly. Video is therefore attractive: monocular reconstruction such as GVHMR~\cite{shenWorldGroundedHumanMotion2024} and PromptHMR~\cite{wangPromptHMRPromptableHuman2025} recovers complex motion from arbitrary footage, at the cost of global drift, temporal jitter and error accumulation that make the result unusable for control without repair.

\paragraph{Learning motion from video.}
FARM~\cite{jingFARMFrameAcceleratedAugmentation2025}, HDMI~\cite{weng2025hdmi} and SONIC~\cite{luo2025sonic} retarget GVHMR or GENMO~\cite{li2025genmo} output to a humanoid; ZEST~\cite{sleiman2026zest} combines MegaSaM \cite{li2025megasam} and TRAM \cite{wang2024tram}. None of them treats reconstruction artifacts systematically. Exbody2~\cite{ji2024exbody2} filters out hard motions with a tuned threshold, while ASAP~\cite{heASAPAligningSimulation2025} and KungfuBot~\cite{xie2025kungfubot} mask or offset physically invalid frames; all of them discard what cannot be made compliant, which is precisely the high-dynamic data we want. Our correction instead repairs the aerial phase.

\paragraph{High-dynamic tracking.}
ASAP adds a delta-action model and fine-tuning; KungfuBot filters via centre of mass and adapts the tracking tolerance to the current error; HuB~\cite{zhangHuBLearningExtreme2025} and Agility Meets Stability~\cite{panAgilityMeetsStability2025} regularise foot and centre-of-mass behaviour. BeyondMimic~\cite{liaoBeyondMimicMotionTracking2025} uses diffusion models and dynamic sampling, and Robot Dancing~\cite{sunRobotDancingResidualActionReinforcement2025} uses residual actions and domain randomisation, but neither sustains long high-dynamic sequences nor handles a fall. OmniXtreme~\cite{wang2026omnixtreme} scales policy capacity and refines for actuation, assuming high-fidelity references.

\paragraph{Recovery.}
HumanUP~\cite{he2025learning} and HoST~\cite{huang2025learning} learn standing-up policies without data, and Unified Humanoid Get-Up~\cite{spraggett2025learning} generalises across morphologies, but all treat recovery as an isolated task. SafeFall~\cite{meng2025safefall} switches to a protective policy, and FIRM~\cite{xu2025unified} unifies prevention, mitigation and get-up from limited demonstrations; BFM-Zero~\cite{li2025bfm} unifies tracking and recovery but needs auxiliary data. StableMimic~\cite{wuStableMimicSmoothHumanLike2026} also unifies tracking and recovery in one policy, and reports smoother, more human-like post-fall motion; it obtains that style by \emph{shaping recovery toward a library of retargeted human get-up references}. Our recovery is shaped by no human reference at all: it is learned from the same tracking objective as the motion itself (Sec.~\ref{sec:unified}), which is what makes it applicable when no paired get-up demonstration exists. None of the works above learns recovery jointly with tracking \emph{from video}, and none does so without human reference data.

\section{KungfuAthlete: Repairing Video-Derived Motion}
\label{sec:dataset}

\subsection{From raw videos to robot trajectories}

The dataset is collected from daily training videos of national-level martial arts athletes: 197 videos, automatically segmented into 1{,}726 sub-clips, each reconstructed with GVHMR~\cite{shenWorldGroundedHumanMotion2024} and retargeted with GMR~\cite{araujoRetargetingMattersGeneral2025}. The corpus mixes ground actions (822 instances) and jumping actions (170 instances); the full per-genre breakdown is given in Appendix~\ref{sec:appendix_dataset}.

\begin{table}[ht!]
\centering
\caption{Kinematic statistics of motion datasets. Entries are means over the whole body, denoted by $w$ in the column headings; frame range is given in frames.}
\label{tab:motion_stats}
\resizebox{\columnwidth}{!}{
\begin{tabular}{lcccl}
\toprule
Dataset  & Body Lin Vel ($w$) & Body Ang Vel ($w$) & Frame Range \\
\midrule
AMASS
& $0.188 \pm 0.165$ 
& $0.510 \pm 0.309$ 
& $[2, 8788]$ \\

PHUMA  
& $0.193 \pm 0.130$ 
& $0.585 \pm 0.368$ 
& $[49, 222]$ \\

LAFAN1  
& $0.790 \pm 0.251$ 
& $1.962 \pm 0.553$ 
& $[3065, 9171]$ \\

KungfuAthlete (Ground)
& $0.496 \pm 0.324$ 
& $1.537 \pm 0.959$ 
& $[24, 15792]$ \\

KungfuAthlete (Jump) 
& $0.948 \pm 0.473$ 
& $2.832 \pm 1.431$ 
& $[25, 2842]$ \\

\bottomrule
\end{tabular}
}
\end{table}

We split the dataset into \textit{KungfuAthlete (Ground)} (no jumping) and \textit{KungfuAthlete (Jump)} (aerial motions). Table~\ref{tab:motion_stats} shows that the Jump subset attains the highest body linear and angular velocities of all compared datasets, while the Ground subset still exceeds low-intensity corpora such as AMASS and PHUMA and sits just below LAFAN1. The corpus therefore spans a wide dynamic range, from sustained ground force production and rapid posture adjustment to highly non-stationary aerial behaviour.

\paragraph{Dataset release.}
\label{sec:dataset_release}
KungfuAthlete is released as follows. \textbf{Scale:} 992 annotated instances (Ground 822 / Jump 170), of which 848 screened samples are publicly released, stored as 30\,fps robot \texttt{qpos} ($3$ position, $4$ quaternion, $29$ DoF). \textbf{Artefacts:} the retargeted G1 reference trajectories used in all experiments; the reconstructed motion as GVHMR predictions in SMPL-H; the segmentation indices and per-clip metadata; the height-correction code and training scripts needed to reproduce every result; and a trained fall-recovery checkpoint. \textbf{Access:} these artefacts are available at the project page given in the abstract; the full dataset will be released publicly under an MIT licence on acceptance. \textbf{Raw video:} the original recordings are not redistributed, as they contain personally identifiable footage; the derived motion is sufficient to use or check the dataset.

\subsection{Repairing root height floating and jitter}
\label{sec:correction}

\begin{figure}[t]
    \centering
    \includegraphics[width=1\linewidth]{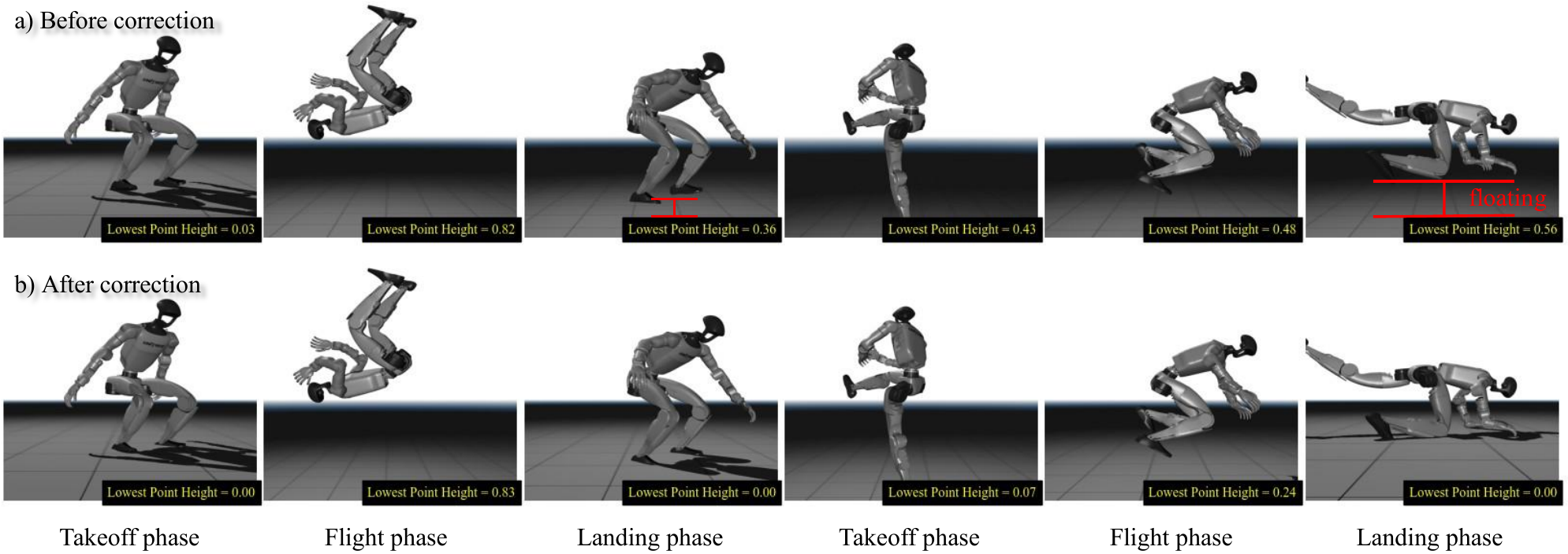}
    \caption{Existing methods exhibit significant height floating when reconstructing jumping motions; prior work (e.g.\ PHUMA) therefore discards such data, wasting jump-related samples (top row). We instead enforce a physically consistent parabolic correction to refine the falling trajectories of jumping motions.}
    \label{fig:corr_height}
    \Description{Comparison of jumping-motion reconstruction with height floating and the proposed physically consistent parabolic correction of the falling trajectory.}
\end{figure}

As shown in Fig.~\ref{fig:corr_height}(a), existing methods~\cite{araujoRetargetingMattersGeneral2025,li2025genmo,shenWorldGroundedHumanMotion2024,lee2025phuma,zhao2026make} often produce a ``floating'' effect during the landing phase when processing jumping motions; as a consequence, prior work (e.g., PHUMA) tends to \emph{discard} such segments entirely, wasting precisely the most dynamic samples a video corpus offers. We instead repair them. Given a retargeted pose sequence $\calQ = \{\mathbf{q}_t\}_{t=1}^T$, where $\mathbf{q}_t = [\mathbf{p}_t, \mathbf{r}_t, \boldsymbol{\theta}_t]$, let $P(t) = p_t^{(z)}$ denote the root height. We enforce ground consistency while preserving temporal trends.

For each frame $t$, let $z_{\min}(\mathbf{q}_t) = \min_{i \in \calB} z_i(\mathbf{q}_t)$ be the world-space height of the lowest body part, so that physically valid poses satisfy $z_{\min} \ge 0$. We then (1) \emph{align the first frame} to the ground, $\hat{P}(1) = P(1) - z_{\min}(\mathbf{q}_1)$; (2) \emph{propagate events}: local extrema of $P(t)$ are found by finite differences, local minima are treated as candidate contacts at which we re-align to the ground provided this introduces no penetration, and elsewhere the height is propagated by truncated velocity integration $\hat{P}(t) = \hat{P}(t-1) + \max(P(t)-P(t-1), 0)$, which preserves upward motion and suppresses downward drift; (3) \emph{interpolate ballistically}: between each local maximum $t_M$ and the following minimum $t_m$ with $\hat{P}(t_M) > \hat{P}(t_m)$, the trajectory follows a constant-acceleration model $\hat{P}(t) = \hat{P}(t_M) - \tfrac{1}{2}g(t-t_M)^2$, whose duration is set by enforcing $\hat{P}(t_m)$; (4) \emph{remove residual penetration} wherever $z_{\min}(\mathbf{q}_t) < 0$; and (5) \emph{smooth} with a Savitzky--Golay filter~\cite{savitzky1964smoothing} to suppress high-frequency jitter while preserving low-order motion structure.

\paragraph{Implementation.}
\label{sec:correction_cost}
The ambiguous local minima of step~(2) are the only manual component; the released code includes the annotation tooling and the height-correction script. Unlike discard-based filtering, step~(3) retains the descending and ascending phases of every detected jump.

\section{Compensating for Missing Actuation with Low-Kinetic-Energy Sampling}
\label{sec:compensate}

\begin{figure}[t]
    \centering
    \includegraphics[width=0.80\textwidth]{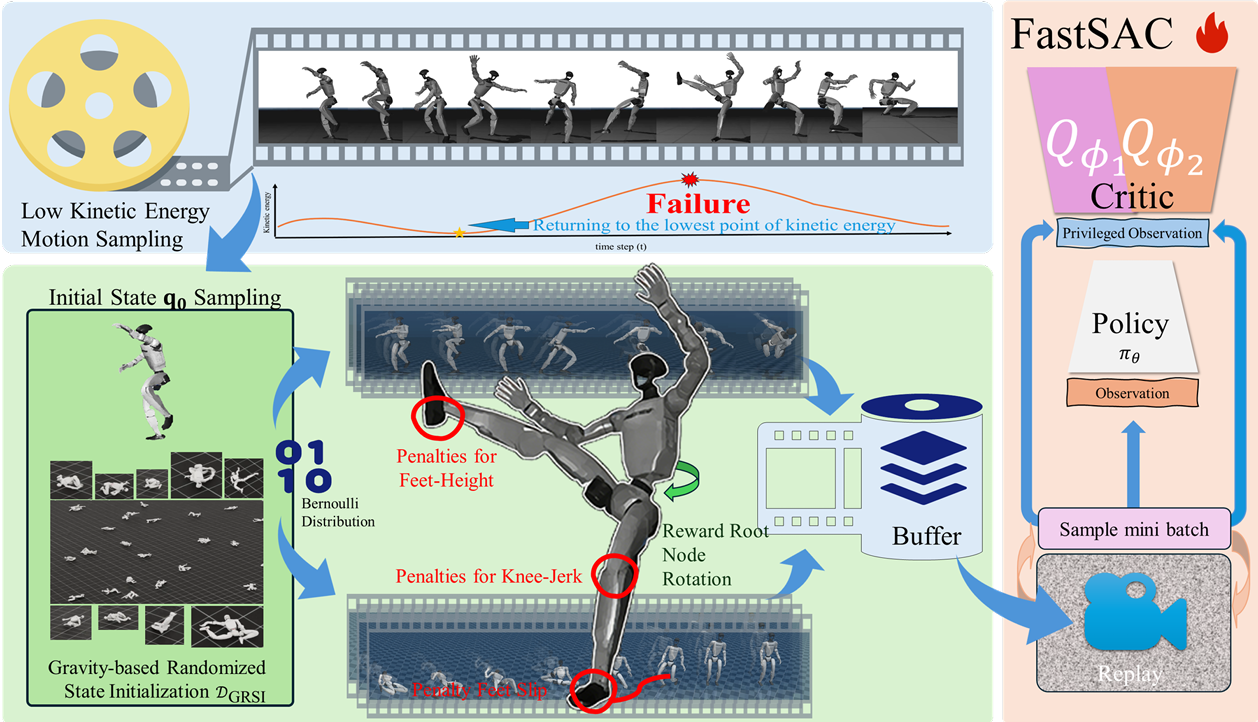}
    \caption{Overview of the training pipeline. \textbf{Top-left:} low-kinetic-energy (LKE) sampling draws video motion from dynamically feasible frames after a tracking failure. \textbf{Bottom-left:} the initial state $\mathbf{q}_0$ is drawn from a Bernoulli mixture over a reference frame and a gravity-based randomized fall state from $\mathcal{D}_{\mathrm{GRSI}}$. \textbf{Centre:} reference-conditioned tracking rewards with recovery penalties (feet height, knee jerk, feet slip, root rotation). \textbf{Right:} the distributional actor--critic loop (FastSAC, Sec.~\ref{sec:setup}).}
    \label{fig:method}
    \Description{Method overview: low-kinetic-energy motion sampling from video, initial-state sampling with gravity-based randomized state initialization, reward terms, and the reinforcement-learning training loop.}
\end{figure}

\subsection{Why strict tracking of video motion fails}

High-dynamic motions reconstructed from videos lack consistent dynamical information and thus violate the robot's intrinsic dynamics constraints. For instance, in highly explosive motions such as backflips, the joint torque distributions (e.g., knee actuation patterns) differ substantially between humans and robots, making strict motion tracking insufficient for learning physically plausible force-generation strategies. This is the concrete manifestation of C2: the reference describes kinematics, whereas the policy must output actuation.

\subsection{Pseudo-low-kinetic-energy (LKE) sampling}
\label{sec:plke}

Because video-captured jumping motions carry no actuation information, the resulting aerial poses may violate the physical feasibility of real robots. To mitigate this, we bias initialisation towards dynamically plausible states while still covering both pre- and post-jump phases. We define a pseudo kinetic energy at frame $t$ as
$E_t = \sum_{i} (v_{t,i}^{\text{joint}})^2
\;+\; \sum_{j} \|\mathbf{v}_{t,j}^{\text{lin}}\|^2
\;+\; \sum_{k} \|\boldsymbol{\omega}_{t,k}^{\text{ang}}\|^2$,
aggregating joint velocities, body linear velocities, and angular velocities.
Let $E_0$ denote the energy of the first frame, and enforce a lower bound to avoid degenerate low-energy bias:
$E'_t = \max(E_t, E_0)$.
Sampling probabilities are then computed via a softmin distribution:
\[
p_t = \frac{\exp(-E'_t)}{\sum_{j=0}^{T-1} \exp(-E'_j)}.
\]

This formulation prioritizes low-energy (i.e., dynamically stable) states while preventing over-concentration on near-static frames, enabling effective learning of jumping behaviours from both take-off and landing contexts.

\textbf{Relation to error-driven sampling.} LKE sampling is the component specific to the video setting, and its defining property is that it is \emph{physics-driven rather than error-driven}: the initial-state distribution is derived from the reference trajectory alone and does not depend on the current policy, so it never concentrates capacity on frames that are dynamically impossible for the robot, however large the tracking error there becomes. Adaptive sampling (as used in BeyondMimic~\cite{liaoBeyondMimicMotionTracking2025}) is instead error-driven: it re-weights initialisation by a kernel whose width follows the tracking error, and can therefore focus on frames that are infeasible precisely because the reference is infeasible. LKE also introduces no kernel-width or decay hyper-parameters to tune. Fig.~\ref{fig:exp_tracking}(b) compares the two.

\subsection{Training pipeline}
\label{sec:curriculum}

The complete reward specification is given in Appendix~\ref{subsec:reward_beyondmimic}: a stage-agnostic tracking reward, terms added in the second stage, and the recovery terms used by the unified policy of Sec.~\ref{sec:unified}.

The training schedule is a standard three-stage curriculum that progressively tightens the tracking-termination tolerances and ends with aggressive domain randomization for sim-to-real transfer; progressive-tolerance stages followed by domain randomization are established practice in high-dynamic humanoid control~\cite{xie2025kungfubot,wang2026omnixtreme,ji2024exbody2}, and we follow that practice here. We adopt the following progressive-tolerance curriculum as a standard training schedule; it is not claimed as a contribution of the paper. High-dynamic motions reconstructed from videos often lack consistent dynamical information and thus violate the intrinsic dynamics constraints of the robot. The curriculum progressively tightens termination conditions to encourage the discovery of force-generation patterns that enable full-motion tracking, rather than mere imitation.

\textbf{Stage I (Exploration).} This stage facilitates comprehensive exploration of the motion space by leveraging diverse sampling strategies (e.g., LKE, adaptive, and uniform), without enforcing strict adherence to the reference motion. Deviations in executing high-dynamic motions (such as jumps and rotations) are permitted, avoiding over-constraining the policy and hindering the discovery of effective behaviours. The reward terms for this stage, adapted from BeyondMimic~\cite{liaoBeyondMimicMotionTracking2025} and augmented with additional DoF terms to encourage more realistic motion, are listed in the ``Stage-agnostic'' block of Table~\ref{tab:rmt}.

\textbf{Stage II (Learning the details).} Building on the framework learned in Stage I, the policy now learns the finer details of the motion under stricter termination conditions and lower tracking-error tolerances. Balance-related and motor-power-related rewards are added on top of the Stage-agnostic terms (Table~\ref{tab:rmt}, ``Stage II additional''). The corresponding termination thresholds are given in Table~\ref{tab:bad_tracking}.

\begin{table}[t]
\centering
\small
\setlength{\tabcolsep}{3pt}
\caption{Bad-tracking termination conditions.}
\begin{tabular}{lllll}
\toprule
Name &  Expression & $\tau_\mathrm{stageI}$ & $\tau_\mathrm{stageII}$ & $\tau_\mathrm{stageIII}$\\
\midrule
Anchor Z Pos. Deviation & 
$\mathbb{I}\Big( \| \mathbf{p}^\mathrm{ref}_{anchor_{z}} - \mathbf{p}^\mathrm{robot}_{anchor_{z}} \|_2 > \tau_\mathrm{pos_{anchor_{z}}} \Big)$ & 0.5 & 0.3 & 0.5\\

Anchor Z Ori. Deviation & 
$\mathbb{I}\Big( | (\mathbf{R}^\mathrm{ref}_{anchor})^\top \mathbf{g} - (\mathbf{R}^\mathrm{robot}_{anchor})^\top \mathbf{g} |_z > \tau_\mathrm{ori_{anchor}} \Big)$ & 0.8 & 0.6  & 0.6 \\

Anchor Ori. Deviation & 
$\mathbb{I}\Big( | (\mathbf{R}^\mathrm{ref}_{anchor_{xy}})^\top \mathbf{g} - (\mathbf{R}^\mathrm{robot}_{anchor_{xy}})^\top \mathbf{g} |_z > \tau_\mathrm{ori} \Big)$ & / & 0.25 & 0.6 \\

Ankle Z Pos. Deviation & $\mathbb{I}\Big( \| \mathbf{p}^\mathrm{ref}_{ankle_{z}} - \mathbf{p}^\mathrm{robot}_{ankle_{z}} \|_2 > \tau_\mathrm{pos_{ankle_{z}}} \Big)$ & 1.25 & 1.0 & 1.25  \\

Wrist Z Pos. Deviation & $\mathbb{I}\Big( \| \mathbf{p}^\mathrm{ref}_{wrist_{z}} - \mathbf{p}^\mathrm{robot}_{wrist_{z}} \|_2 > \tau_\mathrm{pos_{wrist_{z}}} \Big)$ & 0.5 & 0.5  & 0.5 \\

Body Dof Deviation & 
$\mathbb{I}\Big( \exists b \in \mathcal{B}:|\mathbf{p}^\mathrm{ref}_{k,b} - \mathbf{p}^\mathrm{robot}_{k,b} | > \tau_\mathrm{body} \Big)$ & $\pi/2$ & $\pi/8$ & $\pi/6$\\

\bottomrule
\end{tabular}
\label{tab:bad_tracking}
\end{table}

\textbf{Stage III (Aggressive domain randomization).} After Stage II the robot can track the full motion and recover from falls. To close the sim-to-real gap, inspired by OmniXtreme~\cite{wang2026omnixtreme}, we apply more aggressive domain randomization: typical parameters are increased by 50\%, mainly affecting push, initial joint configuration, velocity, gravity, and friction (Table~\ref{tab:dom}). Unlike OmniXtreme, and because of the high dynamic range of our dataset and the difficulty of the jumping task, we additionally randomize robot mass and gravity. Under this aggressive randomization the termination thresholds are relaxed accordingly, allowing the robot to recover from significant deviations and preventing premature termination.

\begin{table}[htbp]
\centering
\caption{Configurations for noise and domain randomization.}
\begin{tabular}{lcc}
\toprule
Parameter Item &  Moderate & Aggressive \\
\midrule
Push Lin. Vel. & $xy$: $\pm 0.5$, $z$: $\pm 0.2$ & $xy$: $\pm 0.75$, $z$: $\pm 0.3$ \\

Push Ang. Vel. & $xy$: $\pm 0.52$, $z$: $\pm 0.78$ & $xy$: $\pm 0.785$, $z$: $\pm 1.17$ \\

Rest Pos. & $/$ & $\pm 0.15$  \\

Rest Lin. Vel.& $/$ & $xy$: $\pm 0.75$, $z$: $\pm 0.3$ \\

Rest Ang. Vel & $/$ & $xy$: $\pm 0.78$, $z$: $\pm 1.17$ \\

Body Mass & 1 & [0.6,1.1] \\

Gravity & -9.81 & [-14.1,-9.81] \\

Dynamic Friction & [0.3,1.2] & [0.2,0.8] \\

\bottomrule
\end{tabular}
\label{tab:dom}
\end{table}

The component specific to our setting is the initial-state distribution of Sec.~\ref{sec:plke}.

\subsection{Positioning: learning high-dynamic skills from unreliable video data}
\label{sec:curriculum_relation}

Table~\ref{tab:prior_compare} contrasts KAB with the closest prior work along three binary axes: whether the reference motion is video-derived, whether a physically invalid aerial phase is \emph{repaired} rather than discarded, and whether initialisation is derived from a kinetic-energy proxy of the reference rather than from the current tracking error. The relevant difference is not the schedule but the reference: because the reference is unreliable, a physics-driven initial-state distribution (Sec.~\ref{sec:plke}) is the appropriate choice.

\begin{table}[t]
\centering
\footnotesize
\setlength{\tabcolsep}{4pt}
\renewcommand{\arraystretch}{1.25}
\caption{\textbf{Positioning on the video-to-robot pathway.} Prior work assumes the processed reference is either learnable or safely discardable; we target the regime where it is neither. \tick{} explicit support, \cross{} absent, `--' not applicable.}
\begin{tabular}{@{}lccc@{}}
\toprule
Method & \shortstack{Video\\ref.} & \shortstack{Aerial\\repair} & \shortstack{Energy\\init.} \\
\midrule
KungfuBot~\cite{xie2025kungfubot}      & \tick  & \cross & \cross \\
OmniXtreme~\cite{wang2026omnixtreme}   & \cross & --     & \cross \\
\midrule
\textbf{KAB (ours)}                     & \tick  & \tick  & \tick \\
\bottomrule
\end{tabular}
\label{tab:prior_compare}
\end{table}

Both prior systems assume that, after processing, the reference is essentially learnable: KungfuBot removes whatever cannot be made physically compliant, and OmniXtreme intervenes on the policy rather than on the data. Neither targets the case where the reference is irreparably wrong in its \emph{most dynamic} phase -- a reconstructed jump that floats and encodes no actuation, which filtering would simply discard. We therefore intervene at three levels at once: repair the reference (Sec.~\ref{sec:correction}), compensate for the missing actuation through physics-driven initialisation (Sec.~\ref{sec:plke}) and a standard curriculum (Sec.~\ref{sec:curriculum}), and unify recovery -- with no get-up reference data of any kind -- so the learned skill survives deployment (Sec.~\ref{sec:unified}).

\section{A Unified Policy for Tracking and Recovery}
\label{sec:unified}

\subsection{Problem formulation}
\label{sec:formulation}

We model video-conditioned control as a reference-conditioned Markov decision process $\mathcal{M}=(\mathcal{S},\mathcal{A},P,r,\gamma)$. At step $k$, the state $s_k\in\mathcal{S}$ concatenates the robot's proprioceptive state (base angular velocity, joint positions and velocities, and the previous action) with the reference state $\mathbf{q}^{\mathrm{ref}}_k$ and its phase index; the action $a_k\in\mathcal{A}\subset\mathbb{R}^{29}$ targets the 29 actuated DoF through joint-level PD controllers; $P(s_{k+1}\mid s_k,a_k)$ is the simulator transition under the domain randomization of Sec.~\ref{sec:curriculum}; and $\gamma$ is the discount factor. The observation and action terms, including the privileged terms available to the critic only, are listed in Appendix~\ref{sec:observation}.

Unlike a motion-capture reference, which records feasible behaviour and makes tracking error a meaningful learning signal, a video-derived reference encodes kinematics without actuation (C2) and may be dynamically impossible in its most dynamic phase (C1). Minimising tracking error alone is thus not well posed, and the remainder of this section replaces it with an objective that rewards matching the reference where possible and returning to a standing, trackable state where it is not.

\subsection{Unified objective}

To make a video-learned skill survive the real world, we formulate autonomous fall recovery and motion tracking as a single task: recovering from arbitrary initial states to a desired target state. We define an end-to-end (one-policy) reward objective:
\begin{equation} 
\max_{\pi}\quad 
\mathbb{E}_{\pi}\!\left[ 
\sum_{k=0}^{\infty}\gamma^k 
\left( 
r_{\mathrm{mt}} + \mathbb{I}_{\mathrm{rc}}(k)\,r_{\mathrm{rc}} 
\right) 
\right],
\label{eq:objective} 
\end{equation}
where $r_{\mathrm{mt}}$ is the physics-based motion-tracking reward and $r_{\mathrm{rc}}$ is the recovery reward (specified in Appendix~\ref{subsec:reward_beyondmimic}), both reference-conditioned functions of the robot state $\mathbf{q}_k$ and reference state $\mathbf{q}^{\mathrm{ref}}_k$. The recovery indicator is
\begin{equation}
\mathbb{I}_{\mathrm{rc}}(k)
=
\mathbb{I}\!\Big(
\big| h_{\mathrm{ref}}(\mathbf{q}_k^{\mathrm{ref}}) - h_{\mathrm{robot}}(\mathbf{q}_k) \big| > \tau
\Big),
\label{eq:recovery_indicator}
\end{equation}
where $h_{\mathrm{ref}}$ and $h_{\mathrm{robot}}$ are the shoulder heights of the reference and the robot at step $k$, and $\tau$ is a threshold on that deviation. Intuitively, when the robot's upper body deviates too far in height from the reference, the task is interpreted as ``recover first'', and the recovery reward activates.

\subsection{State initialization}
\label{sec:init}

The initial state is sampled from a mixture of tracking and recovery conditions:
\begin{equation}
z \sim \mathrm{Bernoulli}(p), \qquad
\mathbf{q}_0 =
\begin{cases} 
\mathbf{q}_0^{\mathrm{ref}}, & z=1,\\[2pt] 
\sim \mathcal{D}_{\mathrm{GRSI}}, & z=0, 
\end{cases}
\label{eq:init}
\end{equation}
where $p$ is the probability of starting from motion tracking and $1-p$ the probability of starting from a fall. The reference is drawn from the dataset, $\mathbf{q}^{\mathrm{ref}}_{0:\infty} \sim \operatorname{Sample}_{\mathrm{LKE}}(\mathcal{D}_{\mathrm{ref}})$, using the LKE sampler of Sec.~\ref{sec:plke}. This decoupling of the objective (Eq.~\ref{eq:objective}) from the state initialization (Eq.~\ref{eq:init}) is what allows a single policy to own both behaviours. We optimise Eq.~\ref{eq:objective} with FastSAC~\cite{seoLearningSimtoRealHumanoid2025}, the algorithm used for every result reported here.

\subsection{Gravity-based randomized state initialization (\texorpdfstring{$\mathcal{D}_{\mathrm{GRSI}}$}{DGRSI})}
\label{sec:grsi}

To enable the robot to recover from various fall poses and resume its tracked motion, we propose a gravity-based method for generating a variety of fall poses. The robot is initialized in a zero-torque mode and released under gravity with randomized contact friction, allowing the system dynamics to naturally generate a wide range of fallen configurations characterised by base position $\mathbf{p}$, base orientation $\mathbf{r}$, and joint configuration $\boldsymbol{\theta}$. The resulting samples form the fall-recovery dataset $\mathcal{D}_{\mathrm{GRSI}}$. No reference motion is associated with these samples; the recovery behaviour is therefore learned entirely from the tracking objective plus the recovery reward.

\subsection{Termination conditions}

At each time step $k$, we define a bad-tracking indicator as the logical disjunction (OR) of all failure conditions of the tracking-termination criterion (listed in Table~\ref{tab:bad_tracking}), yielding $\mathbb{I}_{\mathrm{bad}}(k)$. To avoid premature termination during recovery, we track the length of consecutive bad-tracking steps:
\[
L(k)=
\begin{cases}
L(k-1)+1, & \text{if } \mathbb{I}_{\mathrm{bad}}(k)=1,\\
0, & \text{otherwise}.
\end{cases}
\]
The termination condition is
\[
\mathbb{I}_{\mathrm{bad\_tracking\_terminate}}(k)=
\begin{cases}
\mathbb{I}\!\left(L(k)\ge \tau_{\mathrm{bad}}\right),
& \text{if } \mathbb{I}_{\mathrm{rc}} = 1,\\
\mathbb{I}_{\mathrm{bad}}(k),
& \text{otherwise},
\end{cases}
\]
where $\tau_{\mathrm{bad}}$ is the number of consecutive bad-tracking steps tolerated before termination, set to $150$ steps ($3$\,s at $50$\,Hz).

\section{Experiments}
\label{sec:experiments}

We evaluate KAB in simulation and on real hardware to answer: (Q1)~is the KungfuAthlete dataset sufficiently challenging for representative whole-body controllers? (Q2)~can KAB itself learn high-dynamic motion from this video-derived data, and how does low-kinetic-energy sampling compare with adaptive sampling? (Q3)~does the unified policy provide disturbance rejection and rapid recovery on hardware?

\paragraph{Setup.}
\label{sec:setup}
Experiments run on Ubuntu 24.04 with an NVIDIA RTX 4090. Every result in this paper uses FastSAC~\cite{seoLearningSimtoRealHumanoid2025}, a distributional off-policy actor--critic, with $8{,}192$ parallel environments; the robot is the 29-DoF Unitree G1, and its observation and action terms are listed in Appendix~\ref{sec:observation}.

\subsection{Is the dataset challenging? Stress testing whole-body controllers}
\label{sec:challenging_benchmark}

We select seven representative high-dynamic and complex motions from the dataset: 213 (striking during running), 1480 (hand-supported flip), 278 (backflip), 969 (stands on one leg), 468 (360\textdegree{} side flip), 1307 (long-duration Tai Chi), and 1080 (combining high leg lifts and sweeping kicks). We deploy them on three recent whole-body teleoperation systems -- TWIST~\cite{ze2025twist}, GMT~\cite{chen2025gmt}, and SONIC~\cite{luo2025sonic} -- in MuJoCo. We report mean per-joint position error ($E_{\text{mpjpe}}$, $10^{-3}$\,rad), mean per-joint velocity error ($E_{\text{mpjve}}$, $10^{-3}$\,rad/frame), and success rate over six trials (Table~\ref{tab:teleoperation}).

\begin{table}[htbp]
\centering
\caption{
\textbf{Is the dataset challenging, and can KAB track it?}
$E_{\text{mpjpe}}$ ($10^{-3}$\,rad), $E_{\text{mpjve}}$ ($10^{-3}$\,rad/frame) and success rate over six trials. Motion IDs index clips in the released dataset. The same metrics are reported for KAB, to answer directly whether our own method can track these motions.}
\begin{tabular}{ccccc}
\toprule
Motion ID & Method & $E_{\text{mpjpe}}$ & $E_{\text{mpjve}}$ & succ \\
\midrule
\multirow{4}{*}{78}
    & TWIST & 85.2$\pm$23.8   & 18.9$\pm$9.7   & 0/6 \\
    & GMT   & 68.3$\pm$17.3   & 11.7$\pm$4.8   & 0/6 \\
    & SONIC & 71.9$\pm$17.2   & 4.9$\pm$4.6    & 0/6 \\
    & KAB   & 24.7$\pm$12.1   & 5.4$\pm$4.0    & 6/6 \\
\midrule
\multirow{4}{*}{213}
    & TWIST & 87.4$\pm$21.0   & 21.1$\pm$9.8   & 0/6 \\
    & GMT   & 52.9$\pm$12.8   & 9.7$\pm$5.3    & 0/6 \\
    & SONIC & 78.2$\pm$21.8   & 6.3$\pm$4.4    & 5/6 \\
    & KAB   & 18.5$\pm$7.4    & 4.4$\pm$4.7    & 6/6 \\
\midrule
\multirow{4}{*}{1480}
    & TWIST & 108.3$\pm$54.9  & 18.7$\pm$8.0   & 0/6 \\
    & GMT   & 63.8$\pm$17.6   & 9.2$\pm$3.4    & 0/6 \\
    & SONIC & 167.6$\pm$29.8  & 10.7$\pm$4.7   & 0/6 \\
    & KAB   & 22.3$\pm$16.7   & 5.2$\pm$3.6    & 6/6 \\
\midrule
\multirow{4}{*}{278}
    & TWIST & 100.8$\pm$42.5  & 20.6$\pm$11.4  & 0/6 \\
    & GMT   & 66.8$\pm$26.9   & 11.4$\pm$4.5   & 0/6 \\
    & SONIC & 73.8$\pm$18.2   & 10.0$\pm$6.8   & 0/6 \\
    & KAB   & 30.9$\pm$14.4   & 11.0$\pm$6.5   & 6/6 \\
\midrule
\multirow{4}{*}{969}
    & TWIST & 64.9$\pm$19.3   & 10.6$\pm$7.8   & 0/6 \\
    & GMT   & 55.0$\pm$14.8   & 8.1$\pm$3.3    & 0/6 \\
    & SONIC & 61.8$\pm$5.6    & 2.8$\pm$1.8    & 4/6 \\
    & KAB   & 23.2$\pm$3.7    & 1.2$\pm$1.1    & 6/6 \\
\midrule
\multirow{4}{*}{1307}
    & TWIST & 61.7$\pm$14.6   & 5.7$\pm$3.3    & 0/6 \\
    & GMT   & 63.8$\pm$23.4   & 4.2$\pm$0.5    & 0/6 \\
    & SONIC & 60.7$\pm$11.0   & 1.7$\pm$1.1    & 5/6 \\
    & KAB   & 32.4$\pm$8.0    & 1.5$\pm$1.0    & 6/6 \\
\midrule
\multirow{4}{*}{1080}
    & TWIST & 163.3$\pm$49.1  & 24.4$\pm$13.6  & 0/6 \\
    & GMT   & 96.5$\pm$21.9   & 13.4$\pm$6.0   & 0/6 \\
    & SONIC & 99.5$\pm$33.0   & 10.8$\pm$5.5   & 0/6 \\
    & KAB   & 22.7$\pm$7.8    & 3.6$\pm$3.1    & 6/6 \\
\bottomrule
\end{tabular}
\label{tab:teleoperation}
\end{table}

Existing unified teleoperation systems show clear limitations in handling highly dynamic motions, particularly those involving rapid pose transitions and aerial movement. While SONIC achieves partial success on relatively stable motions, this advantage does not generalise to high-dynamic scenarios, revealing poor robustness under drastic dynamic variations. These findings establish the dataset as a meaningful benchmark and, by the same token, establish that solving it requires the repair-and-compensation pipeline of Secs.~\ref{sec:dataset}--\ref{sec:unified}: KAB, which uses exactly that pipeline, succeeds on all seven motions.

\subsection{Can KAB learn high-dynamic motion from video?}
\label{sec:tracking}

To test whether the pipeline actually learns high-dynamic tracking from video-derived data -- and not merely that the data is hard -- we compare against a single-stage BeyondMimic~\cite{liaoBeyondMimicMotionTracking2025} baseline and quantify accuracy on the full benchmark.

\begin{figure}[htbp]
  \centering
  \begin{subfigure}{\columnwidth}
    \includegraphics[width=\linewidth]{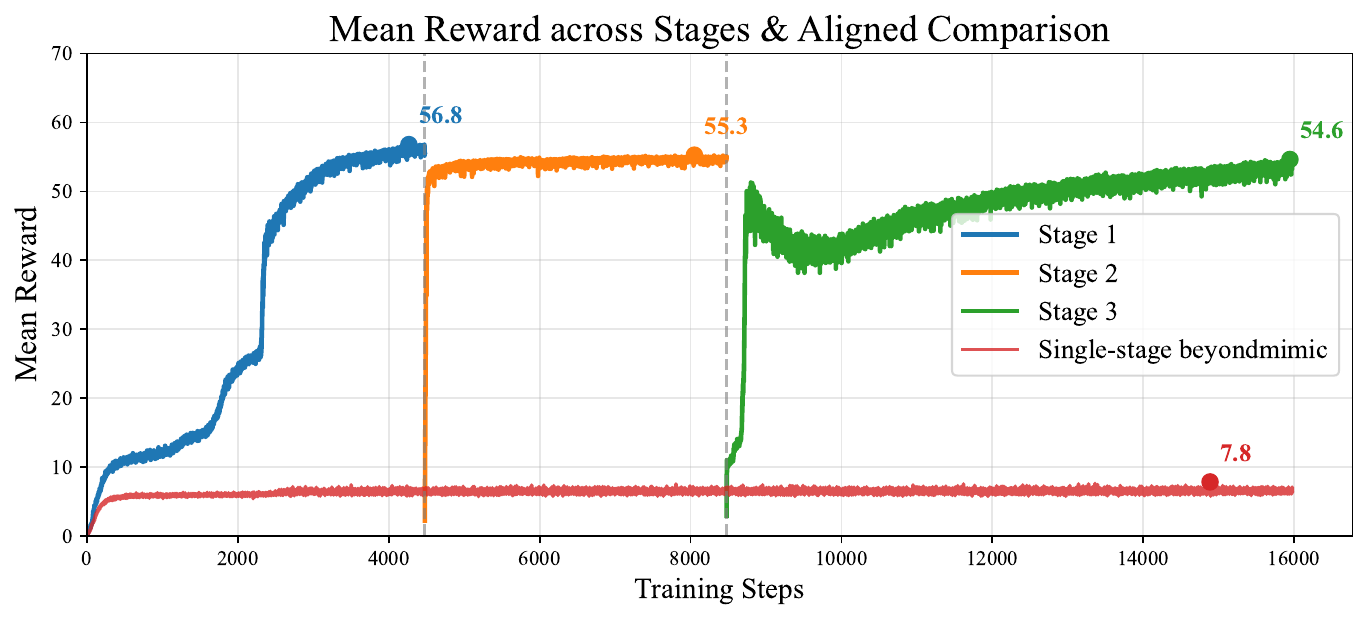}
    \caption{Mean reward variation across the three training stages.}
    \label{fig:reward_threestage}
  \end{subfigure}
  \\[2pt]
  \begin{subfigure}{\columnwidth}
    \includegraphics[width=\linewidth]{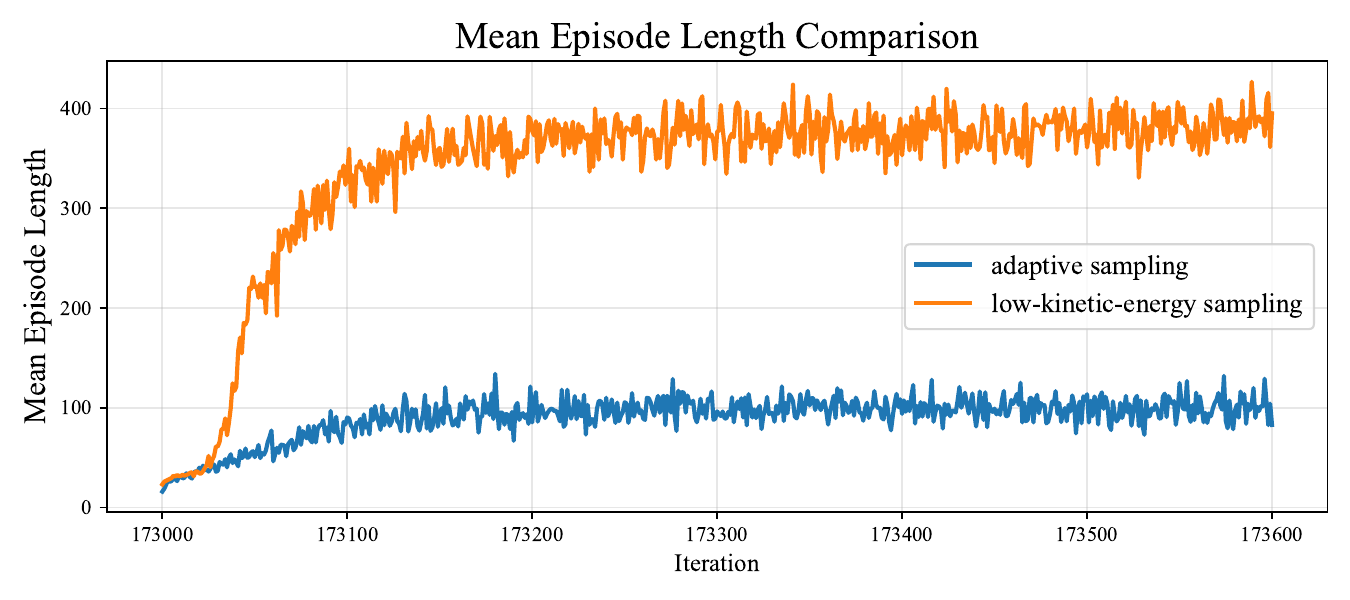}
    \caption{Mean episode length: LKE vs.\ adaptive sampling.}
    \label{fig:compare of lke and adaptive}
  \end{subfigure}
  \caption{
    (a) Reward curves of the three-stage training and the single-stage baseline on Motion 468. (b) Episode length under LKE and adaptive sampling in Stage II on Motion 278. LKE reaches usable tracking behaviour markedly earlier than error-driven adaptive sampling.
    }
  \label{fig:exp_tracking}
    \Description{Training curves comparing three-stage curriculum learning with a single-stage baseline, and mean episode length under low-kinetic-energy and adaptive sampling strategies.}
\end{figure}

\paragraph{Single-stage baseline.}
The single-stage baseline uses identical environment, reward, and termination settings as our Stage II. As shown in Fig.~\ref{fig:reward_threestage}, it cannot track challenging motions such as 468 (360\textdegree{} side flip): early strict tracking constraints cause premature termination and insufficient exploration, trapping the policy in a local optimum with reward stuck at 7.8. Our three-stage curriculum lets the agent first learn the motion's structure and then refine precision, reaching a final reward of 54.6.

\subsection{Sampling strategy: LKE vs.\ adaptive}
\label{sec:lke_ablation}

Fig.~\ref{fig:exp_tracking}(b) compares the two on Motion 278. Error-driven adaptive sampling places its restarts where the current tracking error is largest, which keeps re-initialising the policy from the same infeasible aerial poses it cannot yet realise, so episodes stay short. LKE instead selects states from the physics of the reference and is therefore independent of the policy; this is also why it needs no kernel-width or decay hyper-parameter tuned against a moving error signal. Enlarging BeyondMimic's \texttt{adaptive\_kernel\_size} or \texttt{adaptive\_lambda}~\cite{liaoBeyondMimicMotionTracking2025} does not remove the dependence, since both remain functions of the error of a policy that is still failing.

\subsection{Robustness and rapid recovery}
\label{sec:recovery}

\paragraph{Protocol.}
\label{sec:recovery_protocol}
To make the comparison verifiable we fix one protocol. \textbf{Fall states:} initial states are drawn from $\mathcal{D}_{\mathrm{GRSI}}$ (Sec.~\ref{sec:grsi}) and grouped into four families -- prone, supine, lateral and twisted configurations -- obtained by releasing the zero-torque robot under gravity with randomised contact friction. \textbf{Identical initialisation:} every method starts from the \emph{same} sampled states with a fixed seed per trial. \textbf{Metric:} recovery time is the interval from a fallen state to a stable upright posture (shoulder height within $\epsilon$ of standing and root orientation within tolerance, sustained for 0.5\,s). \textbf{Reported quantities:} recovery time, averaged over the trials in which the method recovers; a method that fails to stand within the horizon is counted as a failure rather than credited with a time, since time alone can favour policies that trade stability for speed. Results average three repeated trials, in simulation and on hardware where a hardware baseline exists.

\begin{table}[htbp]
\centering
\caption{
\textbf{Comparison of fall recovery performance and capability unification.}
Within the methods listed, only KAB combines a reference-free recovery objective with unified tracking and recovery, and KAB attains the shortest recovery time under the protocol above.
}
\label{tab:recovery}

\resizebox{\columnwidth}{!}{
\begin{tabular}{lcccc}
\toprule
Method &
\makecell[c]{Reference-\\free} &
\makecell[c]{Tracking \&\\Recovery} &
\makecell[c]{No Get-Up\\Cmd.} &
\makecell[c]{Recovery\\Time (s)} \\
\midrule
HumanUP~\cite{he2025learning}    & \tick & \cross & \cross & 6.708$\pm$0.987 \\
HoST~\cite{huang2025learning}       & \tick & \cross & \cross & 2.055$\pm$0.310 \\
FIRM~\cite{xu2025unified}       & \cross  & \cross & \tick & 1.729$\pm$0.424 \\
BFM-Zero~\cite{li2025bfm}  & \cross  & \tick & \tick & 1.566$\pm$0.065 \\
Heracles~\cite{tao2026heracles}   & \cross  & \tick & \tick & 2.471$\pm$0.139 \\
StableMimic~\cite{wuStableMimicSmoothHumanLike2026} & \cross & \tick & \tick & n/r \\
\midrule
\textbf{KAB} & \tick & \tick & \tick & \textbf{0.692$\pm$0.136} \\
\bottomrule
\end{tabular}
}
\vspace{3pt}

{\footnotesize \textbf{n/r}: the original paper reports no time-to-standing (recovery is scored as success over 100 matched push-to-fall trials within a 20\,s horizon), and no code, model or video is released, so the quantity cannot be measured under our protocol.}
\end{table}

\paragraph{Results.}
We deploy the learned policy in MuJoCo and on a Unitree G1, using Motion 1307 (a five-minute Tai Chi sequence) as the long-horizon benchmark. Under mild perturbations the robot rapidly stabilises and resumes motion (Fig.~\ref{fig:introduction}(d)); under large disturbances causing falls (Fig.~\ref{fig:introduction}(c)) the policy promptly recovers, stands up, and returns to the target trajectory. Table~\ref{tab:recovery} compares against existing methods: HumanUP and HoST are reference-free but are standing-only policies and cannot track motion; FIRM supports fall buffering and standing but not tracking; BFM-Zero and Heracles unify the two but require recovery data and are slower; StableMimic also unifies the two, but shapes recovery with retargeted human get-up references and does not report a recovery time. KAB unifies tracking and recovery without recovery-specific data at $\approx 0.7$\,s.

\begin{figure}
    \centering
    \includegraphics[width=0.90\linewidth]{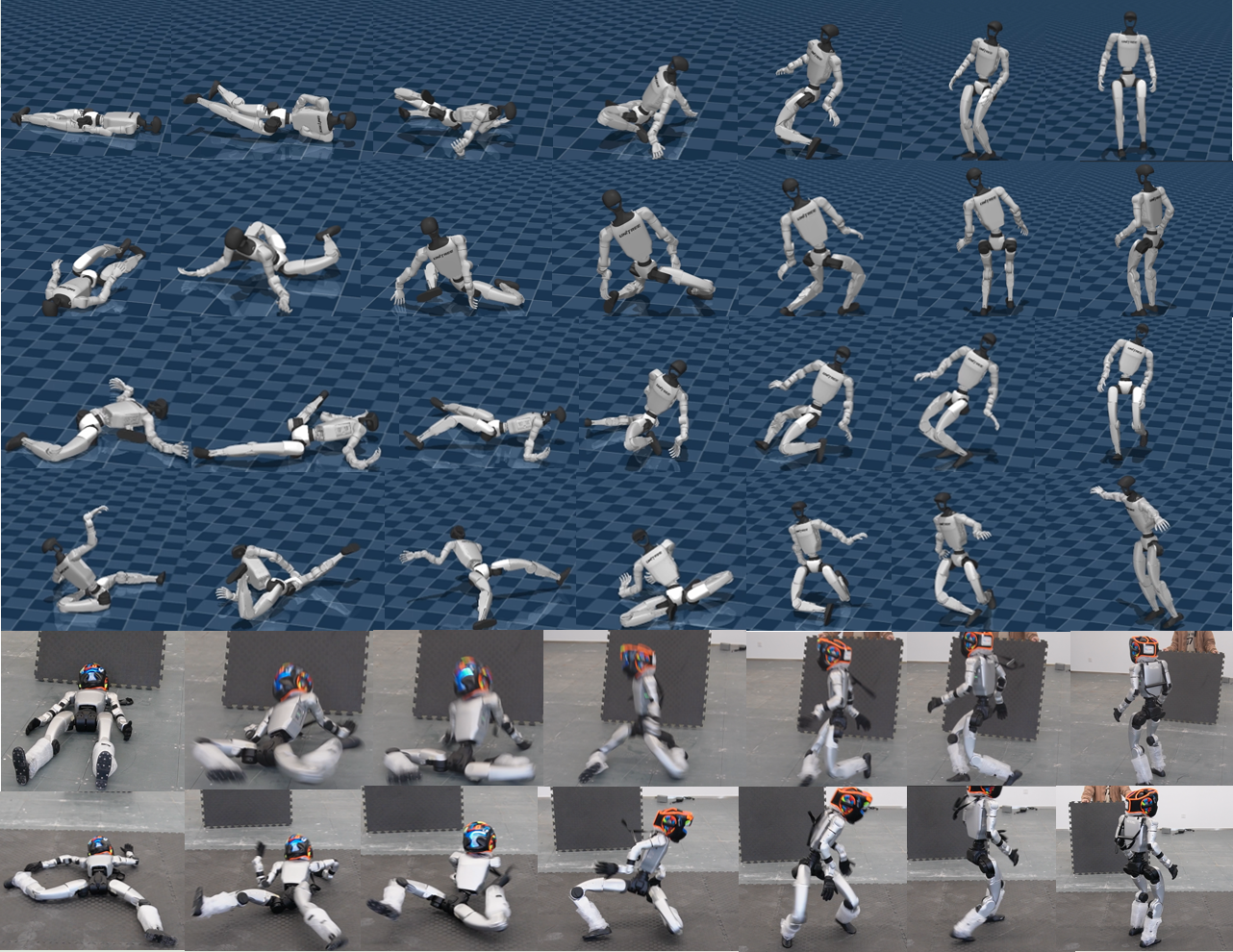}
    \caption{Fall recovery from diverse fallen configurations. Top four rows: MuJoCo recovery sequences covering prone, supine and two twisted joint configurations with irregular limb postures. Bottom two rows: the corresponding real-world recovery on the Unitree G1.}
    \label{fig:recover_sim_real}
    \Description{Fall-recovery sequences from diverse initial fallen configurations in MuJoCo and on the Unitree G1.}
\end{figure}

\paragraph{Real-world evaluation and diverse postures.}
On the physical Unitree G1 the policy recovers from all four fall families and resumes the tracked motion, including the highly distorted twisted configurations whose asymmetric limb arrangements are far from any upright pose; recovery is reproduced in simulation and on hardware (Fig.~\ref{fig:recover_sim_real}), so the policy learns a generalised recovery primitive rather than a fixed one.

\paragraph{Recovery style, without reference data.}
StableMimic~\cite{wuStableMimicSmoothHumanLike2026} reports smoother, more human-like post-fall motion by shaping recovery toward human get-up references, at the cost of requiring a retargeted get-up library. Our recovery uses no human reference at all: it emerges from the unified objective of Sec.~\ref{sec:unified} and is shaped by the robot's own dynamics, which is what makes it applicable when no paired get-up demonstration exists. This is the property reported as reference-free in Table~\ref{tab:recovery}.

\paragraph{Is faster always better?}
\label{sec:fast_recovery}
Minimising recovery time is not unconditionally desirable: an aggressive return to standing increases motor wear, mechanical stress, and the risk to nearby people or objects. What matters is whether the robot becomes usable again \emph{within the task's time budget}, and staged settings make that budget explicit. In the boxing, kickboxing and dance events of the World Humanoid Robot Games~\cite{whrgGames2026} and in RoboCup humanoid soccer~\cite{robocupHumanoidLeagueRules2025}, a robot that is down cannot strike, defend or continue its routine, so time on the ground is time in which the round is lost. Such settings reward a recovery that completes inside the budget, not the fastest possible scramble. Our objective therefore does not optimise time directly: the recovery reward (Appendix~\ref{subsec:reward_beyondmimic}) penalises root motion before standing and rewards smooth, upright completion, so the reported time is an \emph{outcome} rather than a target.

\subsection{Ablations}
\label{sec:ablations}

All ablations use Motion 1307, which requires long-horizon stability and precise tracking.

\begin{figure}[htbp]
  \centering
  \begin{subfigure}{0.48\columnwidth}
    \includegraphics[width=\linewidth]{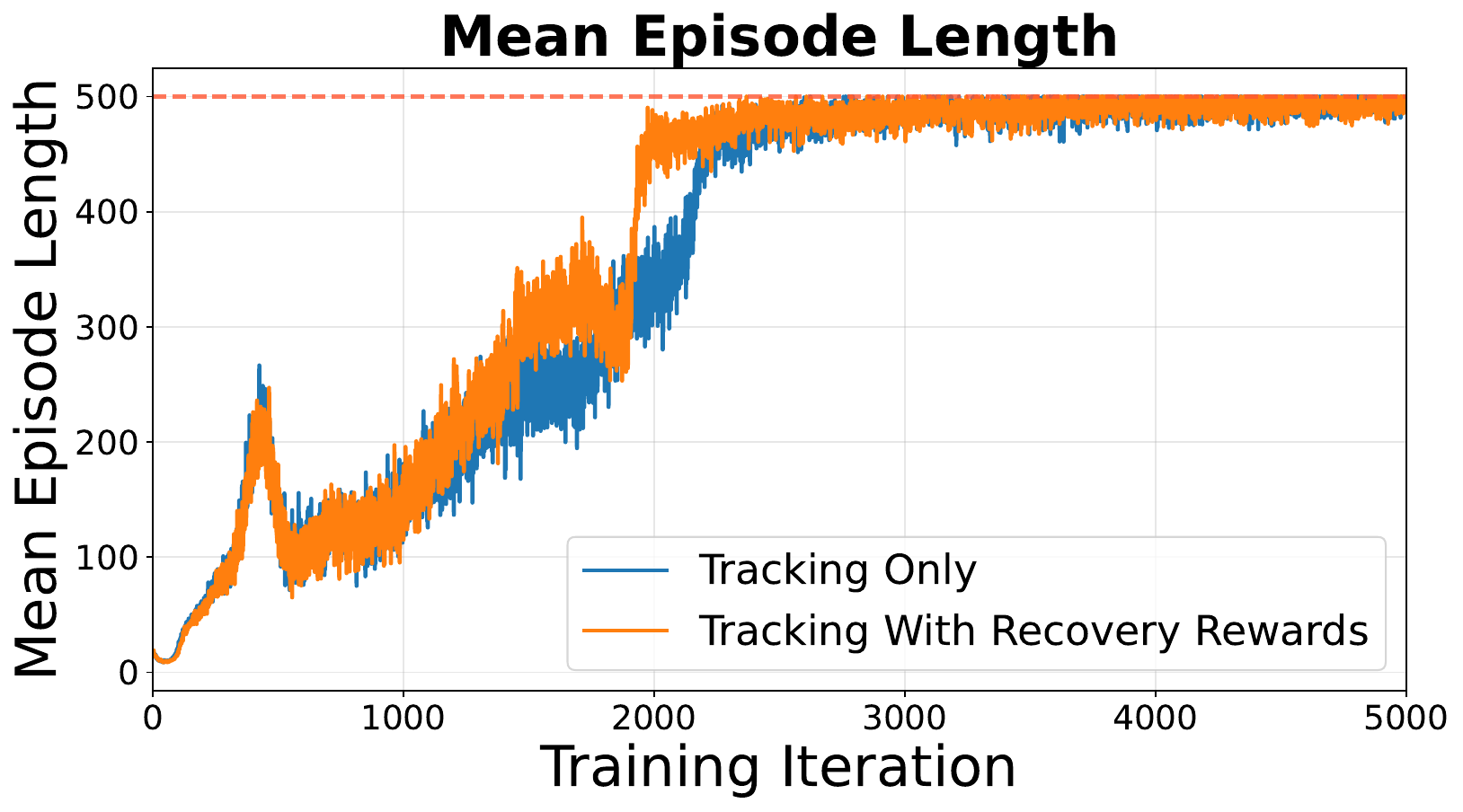}
    \caption{recovery rewards: length}
  \end{subfigure}
  \hfill
  \begin{subfigure}{0.48\columnwidth}
    \includegraphics[width=\linewidth]{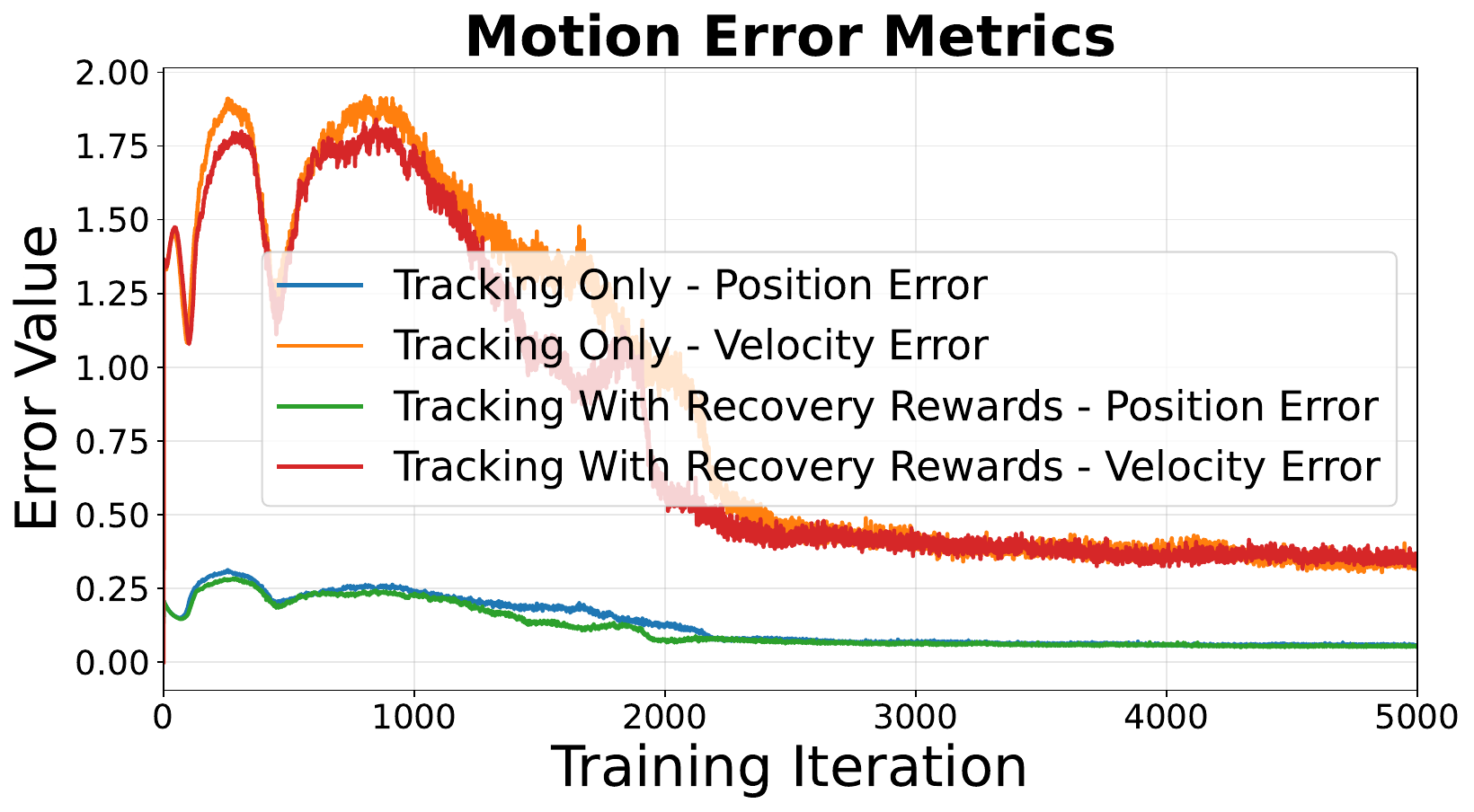}
    \caption{recovery rewards: error}
  \end{subfigure}
  \\[3pt]
  \begin{subfigure}{0.48\columnwidth}
    \includegraphics[width=\linewidth]{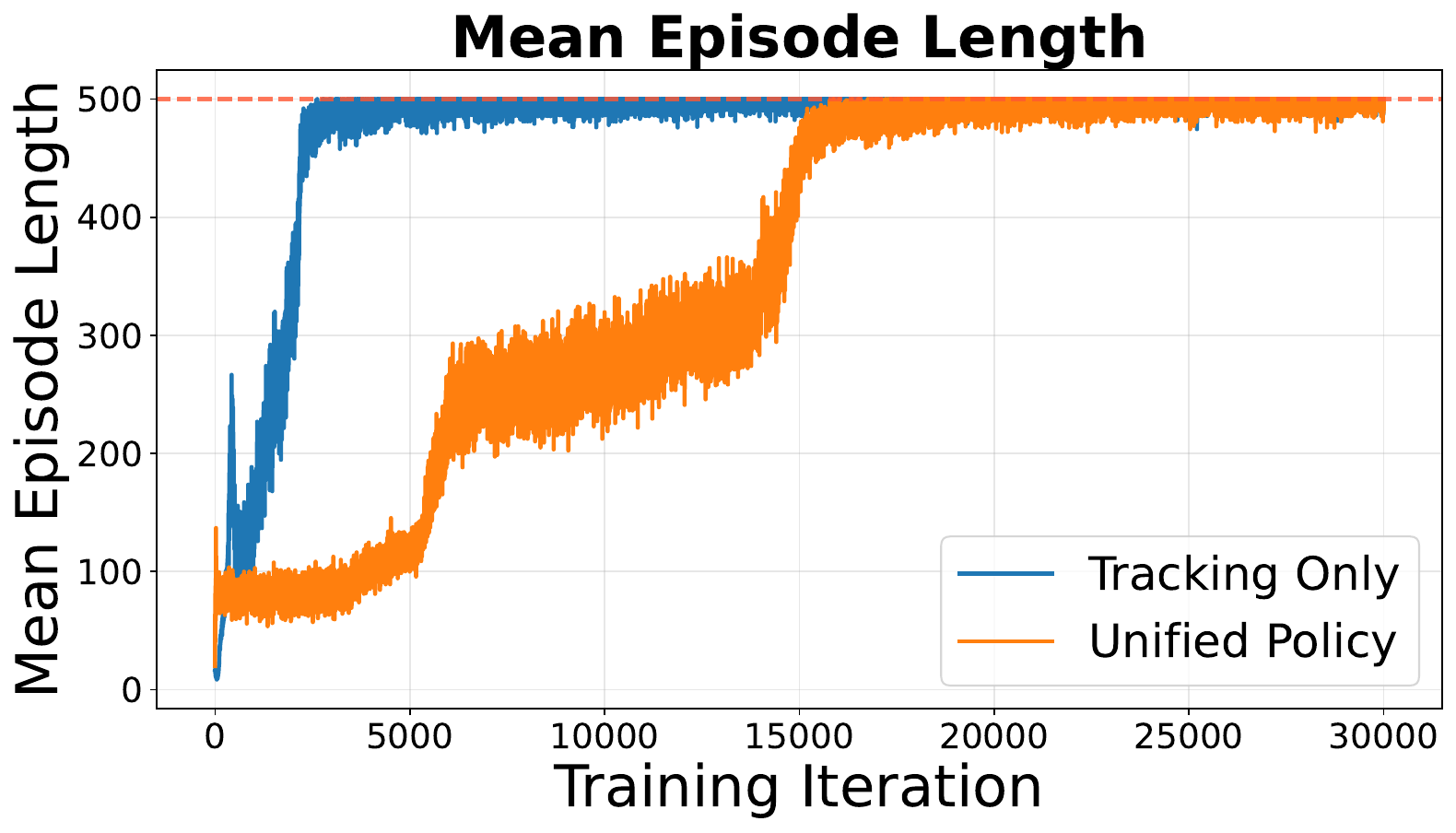}
    \caption{joint design: length}
  \end{subfigure}
  \hfill
  \begin{subfigure}{0.48\columnwidth}
    \includegraphics[width=\linewidth]{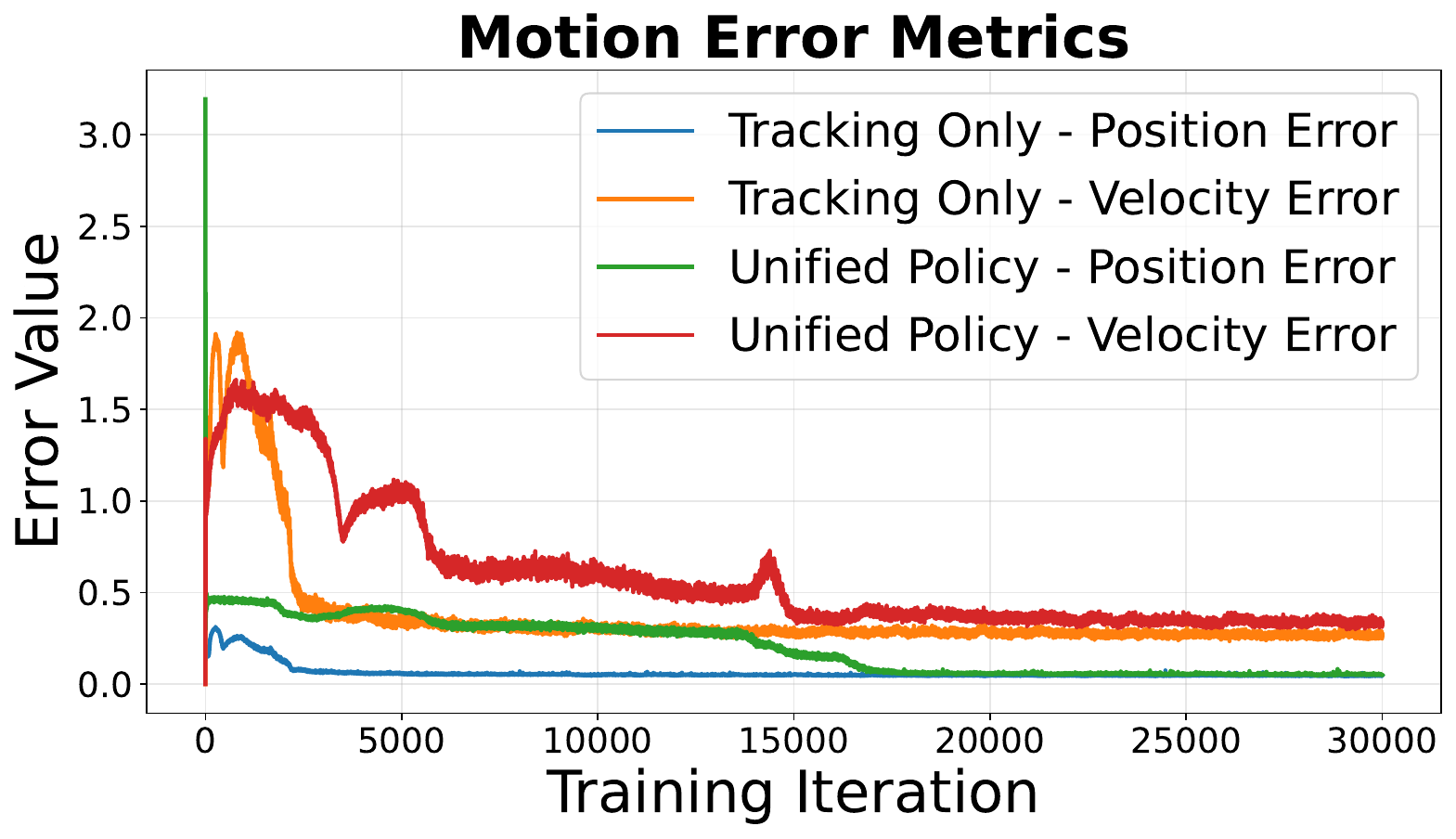}
    \caption{joint design: error}
  \end{subfigure}
  \\[3pt]
  \begin{subfigure}{0.48\columnwidth}
    \includegraphics[width=\linewidth]{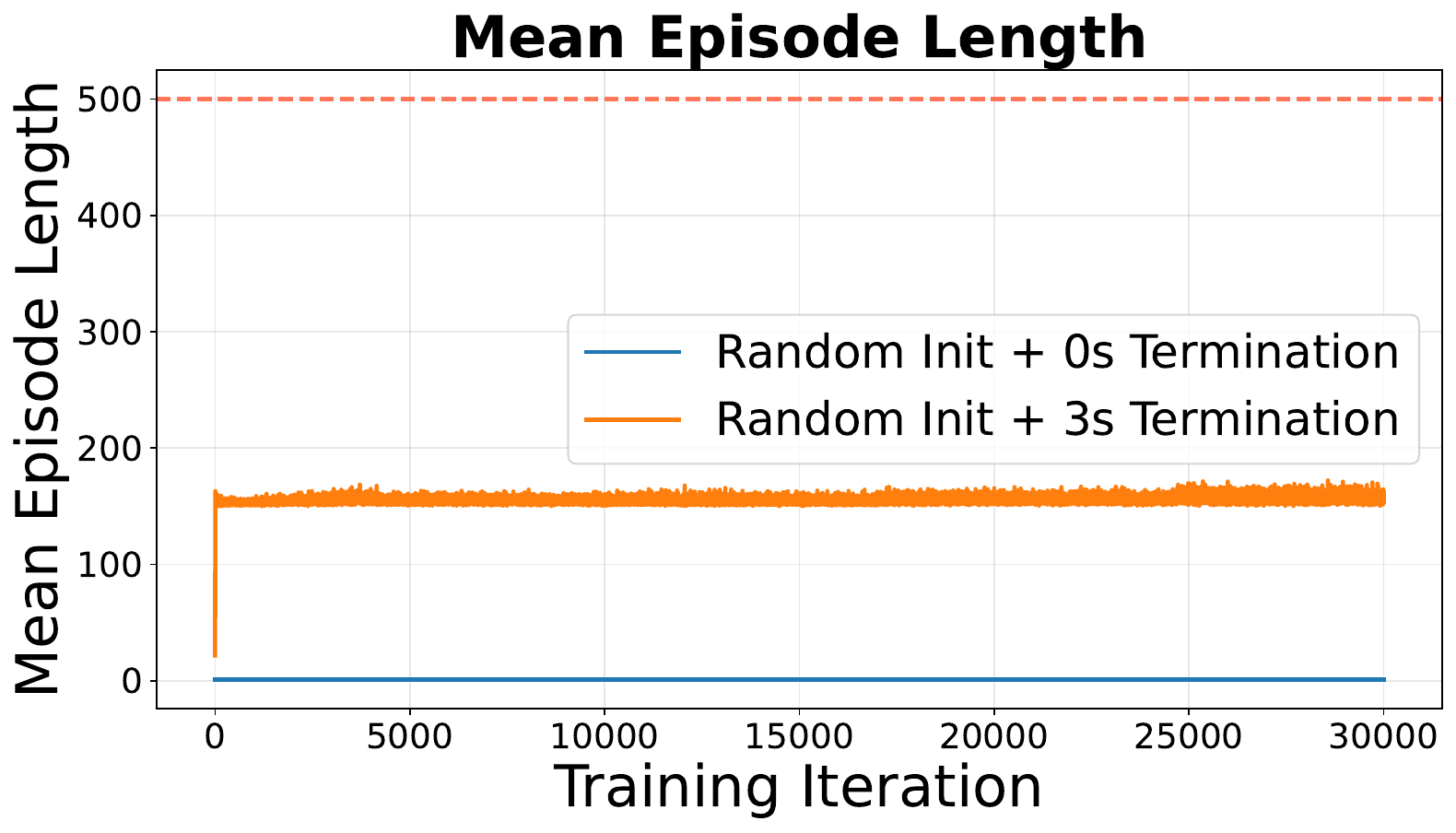}
    \caption{recovery components alone}
  \end{subfigure}
  \hfill
  \begin{subfigure}{0.48\columnwidth}
    \includegraphics[width=\linewidth]{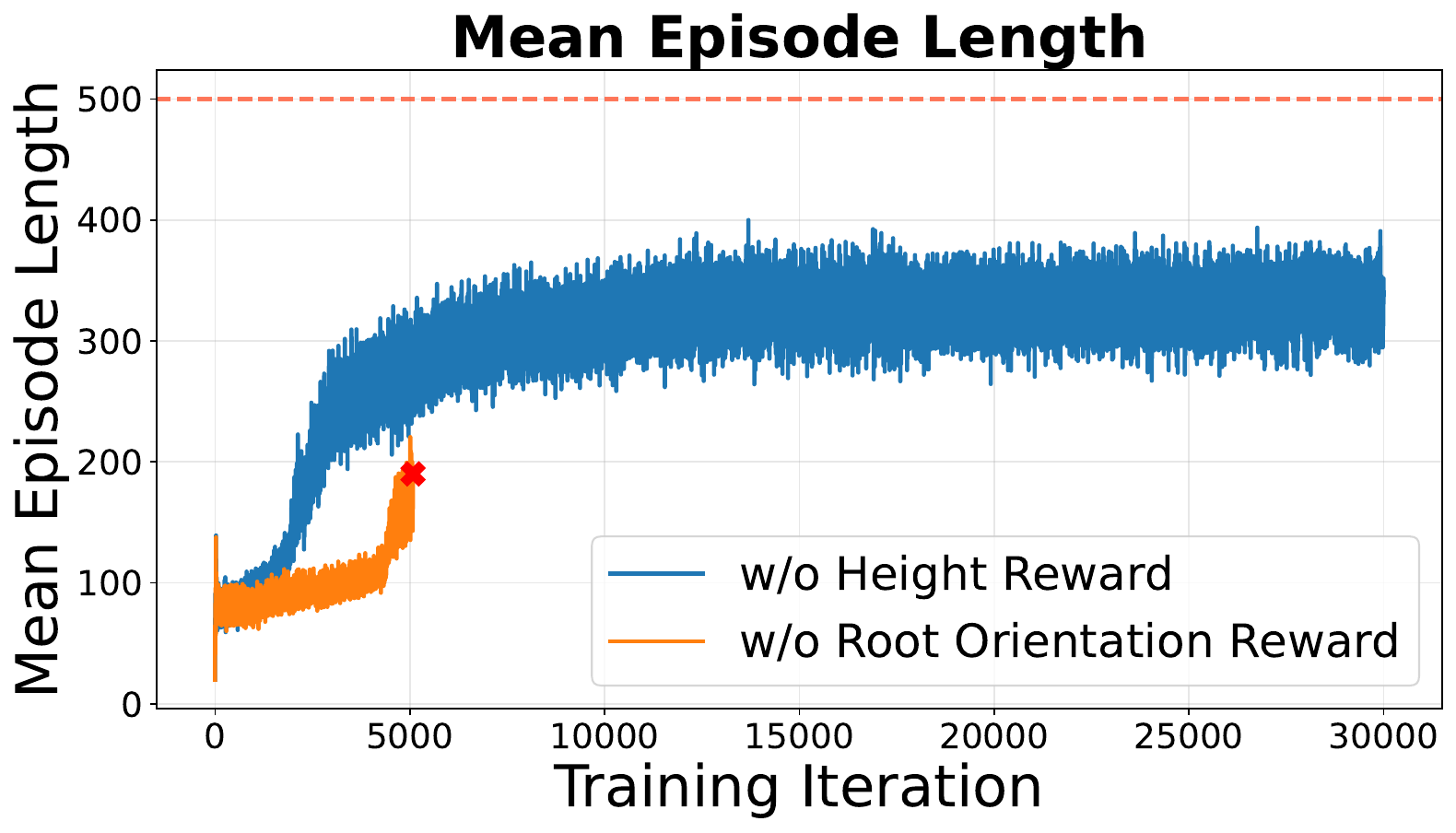}
    \caption{reward-term necessity}
  \end{subfigure}
  \caption{Training curves on Motion 1307. (a,b) Adding the recovery rewards does not change the episode length or the tracking error relative to the tracking-only baseline. (c,d) The full unified configuration matches the baseline's episode length and tracking accuracy, at the cost of slower convergence ($\sim$16k vs.\ $\sim$2.5k iterations). (e) Recovery components without mixed training: with no termination tolerance the episode length stays at 1, and with a 3\,s tolerance it converges to only $\sim$150 of the full 500 steps, so the policy survives but never stands up and tracks. (f) Removing the height reward leaves partial tracking (300--400 steps) without recovery, while removing the root-orientation reward causes early collapse at $\sim$5k iterations.}
  \label{fig:ablations}
    \Description{Ablation curves on Motion 1307 for the recovery rewards, the joint tracking-recovery design, the recovery components alone, and individual reward terms.}
\end{figure}

\paragraph{(i) Recovery rewards do not harm tracking.}
Adding the height and root-orientation rewards leaves the episode-length and error curves essentially unchanged, saturating at the full 500 steps (Fig.~\ref{fig:ablations}a,b).

\paragraph{(ii) Recovery components alone are insufficient.}
With random initial falls and no termination tolerance the episode length stays at 1: the policy is terminated immediately at the start of every episode. With a 3\,s tolerance window (150 steps) it converges to only $\sim$150, i.e.\ the policy survives but neither stands up nor tracks the full motion (Fig.~\ref{fig:ablations}e).

\paragraph{(iii) Joint design works.}
The full configuration reaches the same episode length and a comparable tracking error to the tracking-only baseline, converging at $\sim$16k vs.\ $\sim$2.5k iterations (Fig.~\ref{fig:ablations}c,d).

\paragraph{(iv) Each reward term is necessary.}
Removing the height reward gives partial tracking (episode length 300--400) but no recovery; removing the root-orientation reward destabilises training, which collapses at $\sim$5k iterations (Fig.~\ref{fig:ablations}f).

\section{Conclusion}
\label{sec:conclusion}

We have shown that three failure modes of video-derived motion -- physically inconsistent reconstruction, missing actuation information, and the absence of any failure model -- must be addressed together for the video-to-robot pathway to work. KungfuAthleteBot repairs reconstructed motion with a physics-guided parabolic correction, compensates for missing actuation with LKE sampling and a three-stage curriculum, and unifies tracking with disturbance rejection and fall recovery in a single policy that recovers in $\approx 0.7$\,s and succeeds on every motion in our benchmark, validated on real hardware. Ablations confirm that removing any one component breaks the pipeline. The pathway is not yet end-to-end, since correction still requires manual annotation of ambiguous local minima; scaling to a larger and more diverse motion set, and broadening the evaluation protocols, are natural next steps.

\bibliographystyle{ACM-Reference-Format}
\bibliography{references}


\clearpage
\appendix

\makeatletter
\renewcommand{\@seccntformat}[1]{%
  \@ifundefined{@appendixprefix@#1}{}{\csname @appendixprefix@#1\endcsname}%
  \csname the#1\endcsname\quad}
\newcommand{\@appendixprefix@section}{Appendix~}
\makeatother

\section{Reward Specification}
\label{subsec:reward_beyondmimic}

\begin{table}[t]
\centering
\caption{Physics-stability motion-tracking reward $r_{\text{mt}}$. Rows under ``Stage-agnostic'' are active in \emph{all} stages; rows under ``Stage II additional'' are added on top of them in Stage II. The recovery terms are active whenever the recovery indicator is triggered.}
\small
\setlength{\tabcolsep}{3pt}
\begin{tabular}{lll}
\toprule
Name & Expression & Weight \\
\midrule
\textit{Stage-agnostic tracking reward (Stages I, II, III)} \\[3pt]

Body Dof Match &
$\displaystyle \min \cos\left(\theta_{\mathrm{des}} - \theta\right)$ & 3.0 \\

Penalty Action Rate & 
$\displaystyle \sum_j \| a_j - a_j^\mathrm{prev} \|^2$ & -0.01 \\

Limits DOF Position & 
$\displaystyle \sum_j \left[ (\text{clip}(p_j - \text{upper}_j, 0)) + (\text{clip}(\text{lower}_j - p_j, 0)) \right]$ & -10.0 \\

Penalty Self Collision & $\mathbf{1}_{\text{self-collision}}$ & -10.0 \\

\midrule
\textit{Stage II additional tracking reward} \\[3pt]

Feet Slip & $\displaystyle \sum_f \sqrt{\| \mathbf{v}_f^{xy} \|} \cdot \mathbf{1}_{\text{contact}_f > \text{threshold}}$ & -1.0 \\

Penalty Electrical Power Cost & 
$\displaystyle \sum_{i \in \mathcal{A}} \left( \max\left( -\tau_i \dot{q}_i - 150,\ 0 \right) \big/ 500 \right)^2$ & -10.0 \\

\midrule
\textit{Recovery reward (when $\mathbb{I}_{\mathrm{rc}}=1$)} \\[3pt]

Penalty Relative Shoulder Height & 
$\displaystyle \sum_{b \in \text{shoulder}} \| (\mathbf{p}_b^\mathrm{rel})_z - (\hat{\mathbf{p}}_b)_z \|^2$ & -2.0 \\

Penalty Relative Root Orientation & 
$\displaystyle \| \mathrm{quat\_error}(\mathbf{q}_\mathrm{root}^\mathrm{rel}, \hat{\mathbf{q}}_\mathrm{root}) \|^2$ &  -1.0 \\

Penalty XY Root Movement Before Stand & 
$\displaystyle \mathbf{1}_{\| (\mathbf{p}_\text{shoulder}^\mathrm{rel})_z - (\hat{\mathbf{p}}_\text{shoulder})_z \| > \epsilon} \cdot \| \mathbf{c}_{xy} - \mathbf{c}_{xy}^\mathrm{prev} \|$ & -1.0  \\

\bottomrule
\end{tabular}
\label{tab:rmt}
\end{table}

As shown in Table~\ref{tab:rmt_beyondMimic}, the reward design largely follows that of BeyondMimic, with one key distinction: a stronger emphasis on angular velocity tracking. This encourages the policy to prioritize stable tracking of the robot's global motion during highly dynamic behaviours, rather than overfitting to precise tracking of individual joints.

\begin{table}[t]
\centering
\caption{Motion tracking reward borrowed from BeyondMimic, $r_{\text{mt}}$.}
\begin{tabular}{lll}
\toprule
Name & Expression & Weight \\
Motion Relative Body Position Error& 
$\displaystyle \exp\Bigg(-\frac{1}{\sigma^2} \, \frac{1}{B} \sum_{b} \| \mathbf{p}_b^\mathrm{rel} - \hat{\mathbf{p}}_b \|^2 \Bigg)$ & 1.0 \\

Motion Relative Body Orientation Error & 
$\displaystyle \exp\Bigg(-\frac{1}{\sigma^2} \, \frac{1}{B} \sum_{b} \| \mathrm{quat\_error}(\mathbf{q}_b^\mathrm{rel}, \hat{\mathbf{q}}_b) \|^2 \Bigg)$ & 1.0 \\

Motion Global Body Position Error& 
$\displaystyle \exp\Bigg(-\frac{1}{\sigma^2} \, \frac{1}{B} \sum_{b} \| \mathbf{p}_b^\mathrm{glo} - \hat{\mathbf{p}}_b \|^2 \Bigg)$ & 0.5 \\

Motion Global Body Orientation Error & 
$\displaystyle \exp\Bigg(-\frac{1}{\sigma^2} \, \frac{1}{B} \sum_{b} \| \mathrm{quat\_error}(\mathbf{q}_b^\mathrm{glo}, \hat{\mathbf{q}}_b) \|^2 \Bigg)$ & 1.0 \\

Motion Global Body Angular Velocity& 
$\displaystyle \exp\Bigg(-\frac{1}{\sigma^2} \, \frac{1}{B} \sum_{b} \| \boldsymbol{\omega}_b - \hat{\boldsymbol{\omega}}_b \|^2 \Bigg)$ & 0.4 \\

Motion Global Body Linear Velocity & 
$\displaystyle \exp\Bigg(-\frac{1}{\sigma^2} \, \frac{1}{B} \sum_{b} \| \boldsymbol{l}_b - \hat{\boldsymbol{l}}_b \|^2 \Bigg)$ & 1.0 \\

\bottomrule
\end{tabular}
\label{tab:rmt_beyondMimic}
\end{table}

\section{Observation and Action Specification}
\label{sec:observation}

The released configuration uses the observation terms listed below. Actions are target joint positions for the 29 actuated DoF; actor and critic are three-layer ELU MLPs with hidden sizes $(512,256,128)$, following the \texttt{unitree\_rl\_mjlab} defaults.

\begin{table}[htbp]
\centering
\footnotesize
\caption{\textbf{Observation terms.} Actor and critic share the reference-conditioned and proprioceptive terms; the critic additionally receives privileged terms that are unavailable at deployment. Noise is injected into actor observations only.}
\begin{tabular}{ll}
\toprule
Group & Term \\
\midrule
Actor \& critic & motion command (reference joint targets) \\
                & reference body orientation (anchor frame) \\
                & base angular velocity \\
                & joint positions \\
                & joint velocities \\
                & previous action \\
\midrule
Critic only (privileged) & base linear velocity \\
                & reference body positions (anchor frame) \\
                & robot body positions \\
                & robot body orientations \\
\bottomrule
\end{tabular}
\label{tab:obs}
\end{table}

\section{Dataset Composition}
\label{sec:appendix_dataset}

The corpus comprises ground actions (822 instances: 550 daily training, 93 fist, 99 saber/sword, 80 staff) and jumping actions (170 instances: 74 daily training, 5 fist, 12 saber/sword, 10 staff, 69 skills).

\section{Additional Simulation Results}
\label{sub:sim_motion}

\begin{figure}[t]
    \centering
    \includegraphics[width=\linewidth]{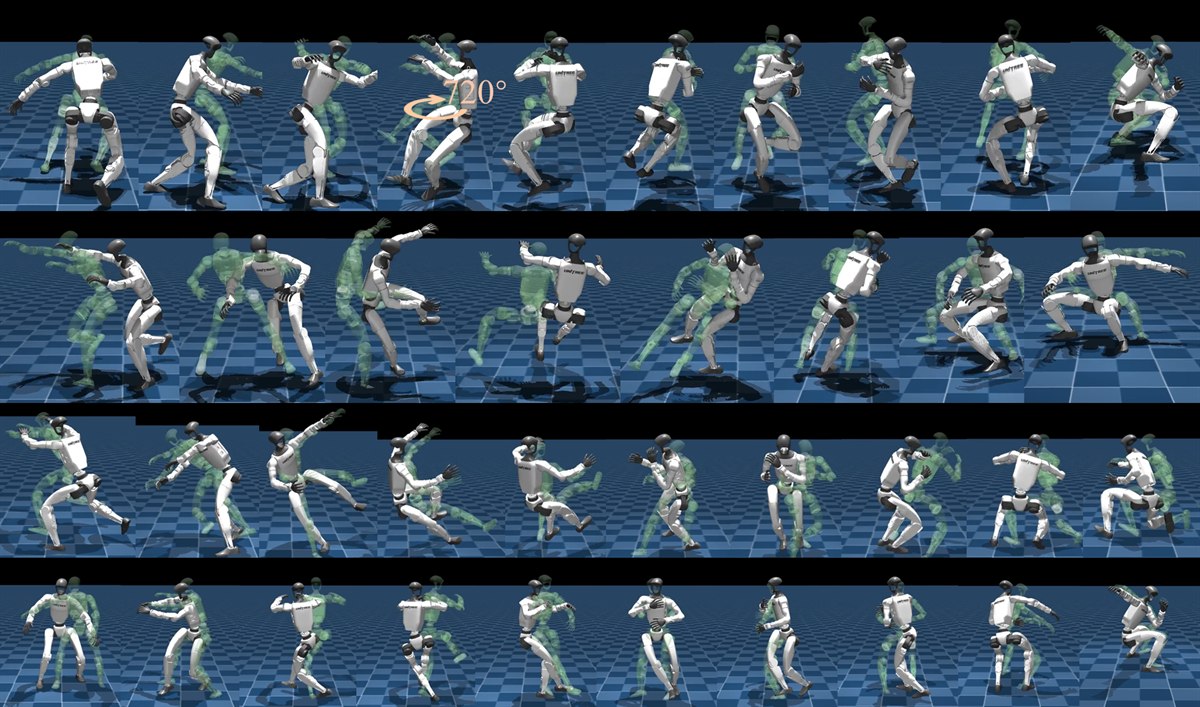}
    \caption{Motion 78: a 720\textdegree{} spin in the air, followed by a 360\textdegree{} kick.}
    \label{fig:78_sim}
    \Description{MuJoCo simulation sequence for Motion 78, showing a 720-degree aerial spin followed by a 360-degree kick.}
\end{figure}

\begin{figure}[t]
    \centering
    \includegraphics[width=\linewidth]{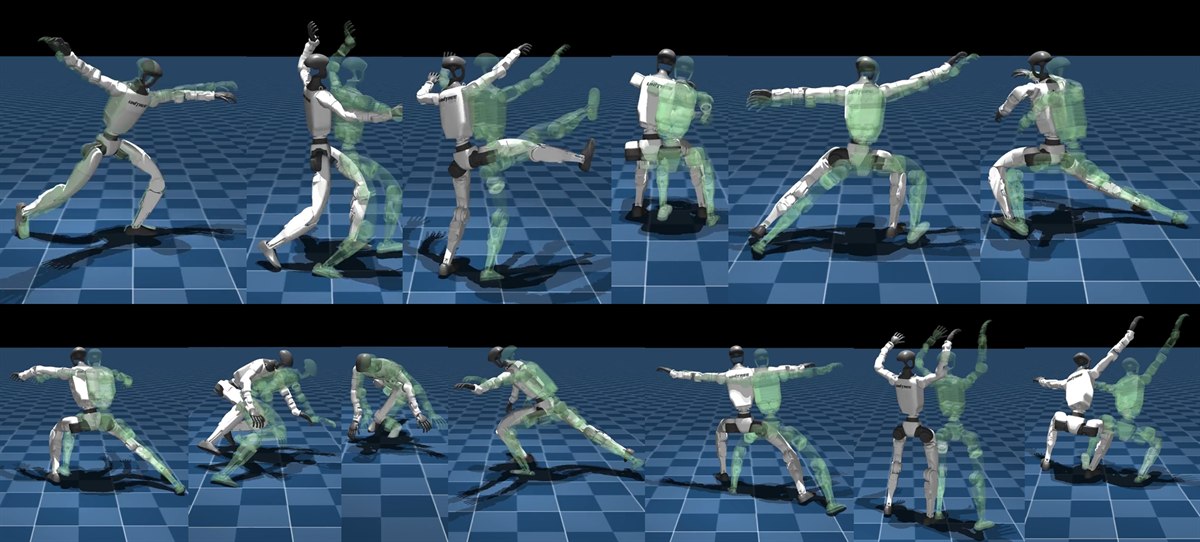}
    \caption{Motion 1080: a kick followed by a roundhouse kick.}
    \label{fig:1080_sim}
    \Description{MuJoCo simulation sequence for Motion 1080, showing a kick followed by a roundhouse kick.}
\end{figure}

\begin{figure}[t]
    \centering
    \includegraphics[width=\linewidth]{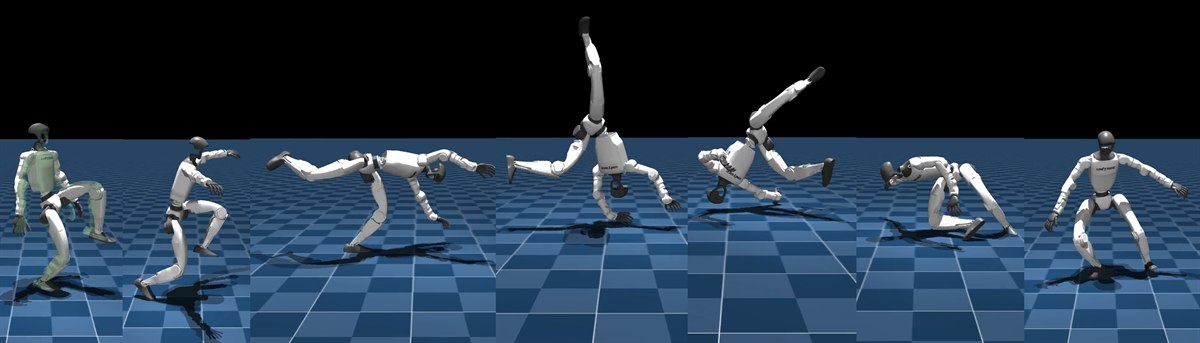}
    \caption{Motion 1480: a side roll.}
    \label{fig:1480_sim}
    \Description{MuJoCo simulation sequence for Motion 1480, showing a highly dynamic side-roll motion.}
\end{figure}

We provide additional MuJoCo simulation results to demonstrate the diversity and
fidelity of the learned high-dynamic motions. Fig.~\ref{fig:78_sim} shows Motion
78 (a 720\textdegree{} aerial spin followed by a 360\textdegree{} kick),
Fig.~\ref{fig:1080_sim} shows Motion 1080 (a kick followed by a roundhouse kick)
and Fig.~\ref{fig:1480_sim} shows Motion 1480 (a side roll); together they cover
strong rotational dynamics, compound striking, and robust ground interaction
under highly dynamic transitions.

\section{Code and Data}
\label{sec:supp_code}

All artefacts are available at the project page \url{\projectpage}. It contains (i)~the retargeting and height-correction code,
(ii)~the training configuration files for all three curriculum stages, (iii)~the
trained fall-recovery checkpoint used in the paper, and (iv)~the released
KungfuAthlete motion data (848 screened samples as 30\,fps robot \texttt{qpos}:
3 position, 4 quaternion, 29 DoF). The dataset, the code and the checkpoint will
be published under an MIT licence on acceptance. The raw athlete videos are not
redistributed, as they contain personally identifiable footage; the derived
motion is sufficient to use or to check the dataset.

\end{document}